\documentclass[10pt,twocolumn,letterpaper]{article}

\usepackage[pagenumbers]{cvpr}      

\usepackage{algorithm}
\usepackage[algo2e]{algorithm2e} 
\usepackage{graphicx}
\usepackage{subcaption}
\usepackage{multirow}
\usepackage{soul}
\usepackage{placeins}
\usepackage[dvipsnames]{xcolor}

\newcommand{\dV}{{dVED}}
\newcommand{\memt}{{MoFM}}

\usepackage{pifont}
\DeclareMathOperator*{\argmax}{argmax} 

\definecolor{cvprblue}{rgb}{0.21,0.49,0.74}
\usepackage[pagebackref,breaklinks,colorlinks,allcolors=cvprblue]{hyperref}

\def\paperID{10620} 
\def\confName{CVPR}
\def\confYear{2025}

\title{MoFM: A Large-Scale Human Motion Foundation Model}

\author{
    Mohammadreza Baharani\\
    The UNC at Charlotte\\
    Charlotte, NC\\
    \texttt{mbaharan@charlotte.edu}\\
    \and
        Ghazal Alinezhad Noghre\\
    The UNC at Charlotte\\
    Charlotte, NC\\
    \texttt{galinezh@charlotte.edu}\\
    \and
    Armin Danesh Pazho\\
    The UNC at Charlotte\\
    Charlotte, NC\\
    \texttt{adaneshp@charlotte.edu}\\
    \and
    Gabriel Maldonado\\
    The UNC at Charlotte\\
    Charlotte, NC\\
    \texttt{gmaldon2@charlotte.edu}\\
    \and
Hamed Tabkhi\\
    The UNC at Charlotte\\
    Charlotte, NC\\
    \texttt{htabkhiv@charlotte.edu}\\}

\begin{document}

\twocolumn[{%
\renewcommand\twocolumn[1][]{#1}%
\maketitle
\begin{center}
    \centering
    \captionsetup{type=figure}
    \includegraphics[width=\textwidth]{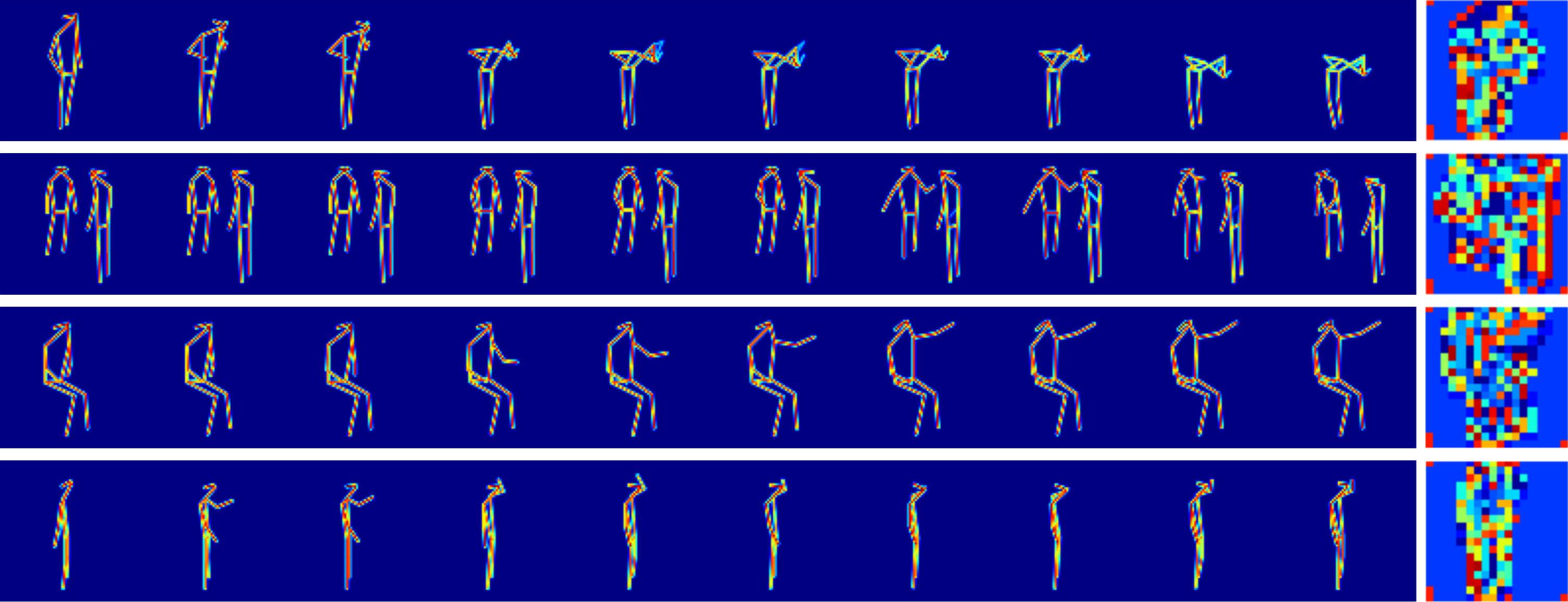}
    \captionof{figure}{Visualization of videos with corresponding pose tokens normalized by vocabulary size. Each row shows skeletal motion frames alongside tokens, with values mapped from blue (low) to red (high), illustrating the alignment of skeletal actions with motion vocabulary.}
    \label{fig:motionbook}
\end{center}%
}]

    

\begin{abstract}

Foundation Models (FM) have increasingly drawn the attention of researchers due to their scalability and generalization across diverse tasks. Inspired by the success of FMs and the principles that have driven advancements in Large Language Models (LLMs), we introduce MoFM as a novel Motion Foundation Model. MoFM is designed for the semantic understanding of complex human motions in both time and space. To facilitate large-scale training, MotionBook, a comprehensive human motion dictionary of discretized motions is designed and employed. MotionBook utilizes Thermal Cubes to capture spatio-temporal motion heatmaps, applying principles from discrete variational models to encode human movements into discrete units for a more efficient and scalable representation. MoFM, trained on a large corpus of motion data, provides a foundational backbone adaptable to diverse downstream tasks, supporting paradigms such as one-shot, unsupervised, and supervised tasks. This versatility makes MoFM well-suited for a wide range of motion-based applications.


\end{abstract}    
\section{Introduction}
\label{sec:intro}

Human-centric computer vision tasks such as pose estimation \cite{sun2019deep, fang2022alphapose, cheng2020higherhrnet}, action recognition \cite{PoseConv3D, geng2023focusing, yang2024one}, and anomaly detection \cite{noghre2024exploratory, hirschorn2023normalizing, wang2023memory} have garnered significant research attention. These tasks are pivotal in various applications, including surveillance systems \cite{pazho2023ancilia, neff2019revamp, ardabili2023understanding} and healthcare monitoring \cite{eswaran2024artificial, khang2024application, luo2023computer}.

Despite substantial progress, effectively capturing the complexity of human motion is a challenging endeavor due to the inherent variability in human poses, dynamic backgrounds, and the high dimensionality of visual data \cite{xie2024dynamic, alinezhad2023understanding}. To tackle these challenges, it is imperative to consider both optimization of the data modalities and development of advanced models capable of effectively understanding these data.

Utilizing the pixel data is a major approach in analyzing videos \cite{wang2023memory, zaheer2022generative, rasheed2023fine}, which can be susceptible to background noise and biases introduced by visual contexts. A common strategy to simplify human motion analysis is to employ pose estimation techniques \cite{sun2019deep, fang2022alphapose, cheng2020higherhrnet}. However, reducing raw pixel data to a sparse set of keypoints can lead to excessive abstraction and loss of valuable information such as fine-grained limb movements and contextual spatial relationships \cite{kocabas2019self, PoseConv3D}. Therefore, a balance must be struck between the high-dimensional raw pixel data and the overly abstracted keypoint representations. To address this issue, using heatmaps as the middle ground between pixels and poses is a well-known practice \cite{feng2023diffpose, luo2021rethinking}. Heatmaps, as the probable location of joints, preserve spatial uncertainty and provide a richer representation of human motion compared to discrete keypoints \cite{PoseConv3D}.

The emergence of Foundation Models (FMs) \cite{wang2024sam, wang2023internimage, chen2024internvl, ma2023crepe} has revolutionized various AI domains, particularly in natural language processing with models like BERT \cite{kenton2019bert}, GPT-3 \cite{brown2020language}, and other Large Language Models (LLMs) \cite{deng2024k2, myers2024foundation}. Such models have demonstrated remarkable scalability and generalization capabilities without extensive task-specific training. These models exhibit exceptional scalability and generalizability, without requiring extensive task-specific training. Consequently, FMs present a promising pathway for advancing human motion understanding.

Inspired by these advancements, we propose a novel Motion Foundation Model (MoFM) designed for the semantic understanding of complex human motions. Instead of processing entire videos or relying solely on keypoint-based poses, MoFM focuses on spatio-temporal heatmaps. To enable large-scale self-supervised training for human motion representation, we introduce a customized discrete Variational Encoder-Decoder (dVED), following principles from discrete Variational Autoencoders (dVAEs) \cite{vahdat2018dvae++, biswas2020dvae, vahdat2018dvae}. Mapping motion heatmaps into Thermal Cubes, we then utilize the proposed dVED to encode them into a discrete latent representation (motion tokens as visualized in \cref{fig:motionbook}). This process creates the MotionBook dictionary, organizing human movements into discrete units. The encoder of dVED functions as a tokenizer to enable BERT-style \cite{kenton2019bert} self-supervised MoFM backbone training through the masking of Thermal Cubes.


The MoFM backbone, trained on a large corpus of data, is designed to serve as a foundational model that can seamlessly adapt to diverse downstream tasks such as action classification and human anomaly detection. Rather than aiming to outperform existing state-of-the-art (SOTA) models, the MoFM backbone’s goal is to provide a flexible, task-agnostic foundation. Through minimal adjustments to the task-specific head, MoFM readily supports a range of applications, from one-shot learning scenarios to unsupervised and supervised tasks, underscoring its versatility for human motion analysis without the need for extensive retraining.

To assess the potential of the MoFM backbone, we target four human-centric downstream tasks: action recognition, one-shot action recognition, self-supervised anomaly detection, and supervised anomaly detection. For each task, we attach a simple, task-specific fully connected head to the MoFM backbone to establish a baseline. The results from this approach demonstrate MoFM’s effectiveness as a foundational model for motion-based applications.

Our experiments highlight its capacity to generalize across tasks, underscoring MoFM’s versatility as a comprehensive solution for diverse human motion understanding. The contributions of this paper are:

\begin{itemize}
    \item A discrete Variational Encoder-Decoder (dVED) designed to encode Thermal Cubes into a discrete latent space (MotionBook) for structured human movement representation.
    \item Pose-aware self-supervised BERT-style training to apply masked Thermal Cubes for task-agnostic learning of human motion.
    \item Motion Foundation Model (MoFM) for semantic understanding of human motion, serving as foundational backbone for diverse downstream tasks.
\end{itemize}
\section{Related Works}
\label{sec:related_works}

Various approaches have been proposed to address the challenges of understanding human motion, including pixel-based \cite{bertasius2021space, wang2023memory, chen2024internvl}, keypoint/graph-based \cite{yan2018spatial, shi2019two, yang2021unik}, and heatmap-based methods \cite{PoseConv3D, feng2023diffpose}. Beyond these modalities, researchers have explored diverse model architectures to leverage them effectively, such as one-shot CNN-based methods \cite{Zhou2017, Pavlakos2018, sarandi2021metrabs}, Graph Convolutional Networks (GCNs), and Transformers \cite{LinWang2021, PoseFormer, Xu2021, Wehrbein2021, PoseSeq2Seq, 10030830}.

Building on these representations, discrete Variational Autoencoders (dVAEs) \cite{RameshZSdVAE, VqVAE, VqVAE2, jang2017categorical} have been developed to capture low-frequency image structures by constructing a quantized latent space. This latent encoding has enabled the use of dVAE encoders for training transformers to learn joint distributions between text and images \cite{RameshZSdVAE} and to model spatial correlations between image patches \cite{dosovitskiy2021an, bao2022beit}. These self-supervised transformers can then be fine-tuned for tasks such as image classification and object localization.

Building on these advancements, our work leverages the strengths of foundational models and discrete representation learning to address the unique challenges in human motion understanding. By combining the generalization of transformer with the structured representations of discrete variational encoders, our Motion Foundation Model (MoFM) introduces a self-supervised approach that captures both the spatio-temporal granularity of motion heatmaps and the interpretability of discrete motion tokens. This approach enables MoFM to generalize effectively across various human motion tasks, establishing it as a versatile backbone for applications such as action recognition and anomaly detection.

Foundation models have revolutionized natural language processing (NLP) and computer vision through large-scale pretraining and self-supervised learning, achieving remarkable generalization capabilities and enabling easy fine-tuning for domain-specific tasks \cite{Devlin2019BERTPO, brown2020language, myers2024foundation}. In NLP, models such as BERT \cite{Devlin2019BERTPO}, GPT-3 \cite{brown2020language}, and K2 \cite{deng2024k2} have demonstrated the effectiveness of training on unlabeled data to capture context and language nuances. Furthermore, transformer architectures with attention mechanisms \cite{vaswani2017attention, dosovitskiy2021an} have been integral to the success of foundational models across tasks.

In computer vision, foundation models \cite{kirillov2023segment, chen2022internvl} have adapted these principles to address complex visual tasks. They leverage masked image modeling \cite{bao2021beit}, vision transformers \cite{dosovitskiy2021an}, and extensive datasets to learn meaningful visual representations.

\begin{figure}[t]
    \centering
    \begin{subfigure}[b]{\linewidth}
        \includegraphics[width=\linewidth]{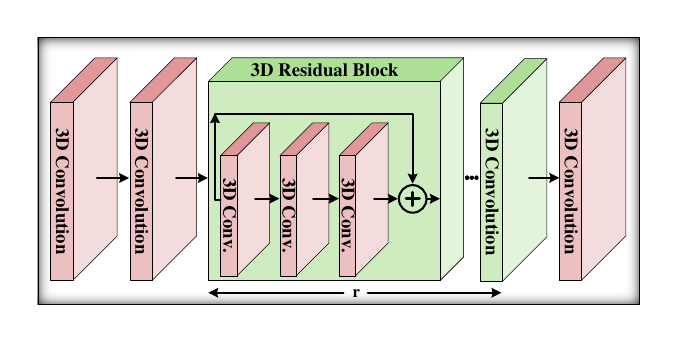}
        \caption{dVED Encoder.}
        \label{fig:dvede}
    \end{subfigure}
    \\
    \begin{subfigure}[b]{\linewidth}
        \includegraphics[width=\linewidth]{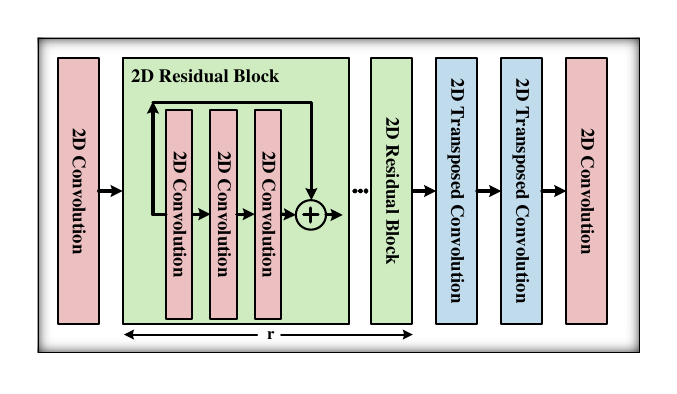}
        \caption{dVED Decoder.}
        \label{fig:dvedd}
    \end{subfigure}
    \caption{Architecture of the proposed custom dVED encoder and decoder. The variable \textit{r} represents the number of layers in which the 2D/3D residual block can be repeated.}
    \label{fig:dved}
\end{figure}
\begin{figure*}[h]
    \centering
    \begin{subfigure}[b]{\textwidth}
        \includegraphics[width=\textwidth]{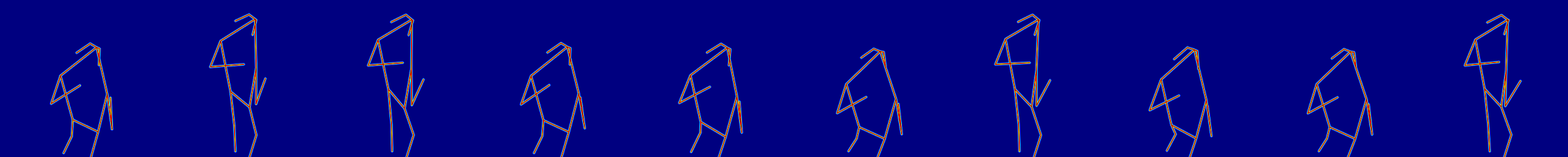}
        \caption{Ground truth heatmap skeleton}
        \label{fig:gt}
    \end{subfigure}
    \\
    \begin{subfigure}[b]{\textwidth}
        \includegraphics[width=\textwidth]{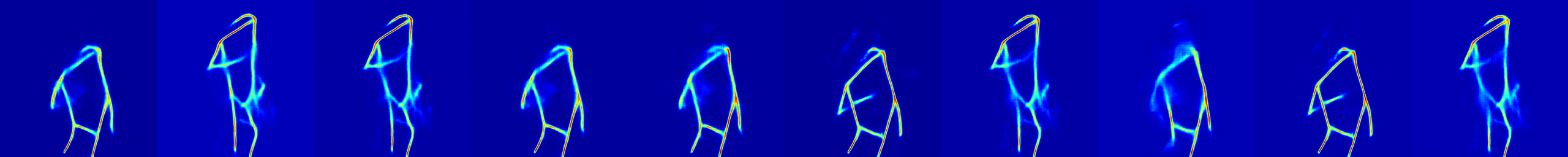}
        \caption{Reconstructed heatmap skeleton.}
        \label{fig:dvae}
    \end{subfigure}
    \caption{Comparison of heatmap skeletons: (a) Ground truth heatmap skeleton used as input for \dV; (b) Reconstructed heatmap skeleton generated by \dV. A ghosting effect is observed for moving joints in the \dV~ output.}
    \label{fig:combined_frames}
\end{figure*}

\section{Problem Formulation}
The objective of this work is to develop a robust foundational model, MoFM, that learns the semantics of human motion using representations called thermal cubes and pose tokens, enabling an efficient and generalized understanding of human motion for downstream tasks. This process begins by constructing a spatio-temporal heatmap, \( U \), for a sequence of \( F \) frames in the 2D pose skeleton \( P = \{P_i\}_{i=0}^{F-1} \). Subsequently, \( U \) is divided into Thermal Cubes, denoted as \( C = \{C_i\}_{i=0}^{K-1} \), where \( K \) represents the number of tokens MoFM backbone can process simultaneously. 

To facilitate self-supervised training, we first employ the Thermal Cubes \( C \) to train a dVED, generating a ``MotionBook" vocabulary \(\mathcal{V}\). With the vocabulary size configured as the transformer head (T-Head) dimension and applying pose-aware BERT-style masked modeling, we train the task-agnostic MoFM backbone to capture motion semantics across tokens. Subsequently, the pretrained MoFM backbone serves as encoder for downstream applications. \cref{fig:self_trans} illustrates this framework, with detailed steps provided in the following sections.

\section{MoFM}
\label{sec:approch}

\subsection{Spatio-Temporal Pose Representation}
Given an input sequence of 2D skeleton keypoints across $F$ frames for $J$ joints,  our initial step involves creating a spatio-temporal heatmap. This is achieved by implementing $J$ Gaussian functions across the $F$ frames as follows:
\begin{align}
U_f=e^{-\frac{(i-x_{jf})^2+(i-y_{jf})^2}{2\times\sigma^2}}, \label{eq:guss_factor}
\end{align}
where $\sigma$ is responsible for determining the spread of the energy around each joint, labeled as $j$, located at coordinates $\langle x_{jf}, y_{jf}\rangle$ for the frame $f$.  A max-pooling filter, spanning over \( J \) joints, is applied to the thermal energy matrix \( U | U \in \mathbb{R}^{J \times F \times H \times W} \), where H and W are the height and the width of the heatmap, respectively. This operation condenses \( U \) into a matrix of reduced dimensions \( U_R \), where \( U_R \in \mathbb{R}^{F \times H \times W} \). This technique simplifies the representation of thermal energy across frames by emphasizing the most significant energy values in each region of the heatmap. Both \( U \) and \( U_R \) are subsequently utilized for training the dVED.

\subsubsection{Thermal Cubes}
To configure the data for processing by a transformer (MoFM backbone), the thermal matrix \(U\) must be segmented into multiple cubes, capturing both the spatial and temporal dimensions of the motion. Thus, \(U\) is transformed into \(K = \frac{HW}{D_S \times D_S}\) cubes, where each cube \(C_i \in \mathbb{R}^{J \times F \times D_S^2}\) and $D_s$ is the patch size. In this work, we centered and scaled the skeleton to fit a heatmap size of \(H = W = 72\). The skeleton structure has \(J = 17\) joints, and \(F = 48\). We also set \(D_S\) to 4, resulting in the thermal matrix \(U\) being divided into \(K = 324\) cubes, each with the shape \(C_i \in \mathbb{R}^{17 \times 48 \times 4 \times 4}\).

\subsubsection{Motion Tokens}\label{subsec:pose_token}
To enable masked input self-supervised training of a transformer, we first need to convert thermal cubes into discrete tokens. Following the methodology outlined by \cite{RameshZSdVAE, jang2017categorical}, we employ a custom discrete Variational Encoder-Decoder (dVED) to facilitate the mapping function \( \mathcal{G}: U \to \widehat{U}_{R} \):
\begin{align}
    \widehat{U}_{R} = f_{\text{dec}}(f_{\text{GS}}(f_{\text{enc}}(U; \theta), \tau); \phi),
\end{align}
where \( \theta \) and \( \phi \) represent the encoder and decoder learnable parameters, respectively, and \( \tau \) is the Gumbel-Softmax temperature parameter. The encoder, Gumbel-Softmax approximation, and decoder are represented by \( f_{\text{enc}} \), \( f_{\text{GS}} \), and \( f_{\text{dec}} \), respectively. 
Formally, the encoder component of the dVED, which functions as a tokenizer, converts the initial heatmap, \( C = \{C_i\}_{i=0}^{K-1} \), into \(K\) discrete tokens, denoted as \( z = \{z_i\}_{i=0}^{K-1} \). Each \(z_i\) is a categorical variable with class probabilities \( \pi = \{\pi_i\}_{i=0}^{T-1} \). \( T \) indicates the size of the vocabulary. Using the Gumbel-Softmax trick, we have: 
\begin{align}
    z_i = \text{one\_hot}(\argmax_i{[g_i + \log\pi_i]}),
\end{align}
where \( g = \{g_i\}_{i=0}^{T-1} \) are samples drawn from Gumbel(0, 1)\footnote{The random variable \( g \) distributed as \( \text{Gumbel}(0, 1) \) can be generated by \( g = -\log(-\log(u)) \), where \( u \) is drawn from a uniform distribution, \( u \sim U(0, 1) \). Alternatively, it is also common to use exponential distribution sampling, where \( g = -\log(x) \), and \( x \sim \text{Exp}(\lambda) \) is drawn from an exponential distribution with a rate parameter \( \lambda = 1 \).}. 
The vocabulary, \(\mathcal{V}\), is constructed as a MotionBook and is defined by the following matrix:
\begin{align}
\mathcal{V} = 
\begin{pmatrix}
v_{0,0} & v_{1,1} & \cdots & v_{0,D-1} \\
\vdots & \vdots & \ddots & \vdots \\
v_{T-1,0} & v_{T-1,1} & \cdots & v_{T-1,D-1} \\
\end{pmatrix},
\end{align}
where \( D \) denotes the dimensionality of each vocabulary entry. The motion dictionaries are trained and later used by the decoder to reconstruct \(\widehat{U}_R\). 
\begin{figure*}[h]
  \centering
  \includegraphics[width=1.0\textwidth,,trim= 10 10 20 10,clip]{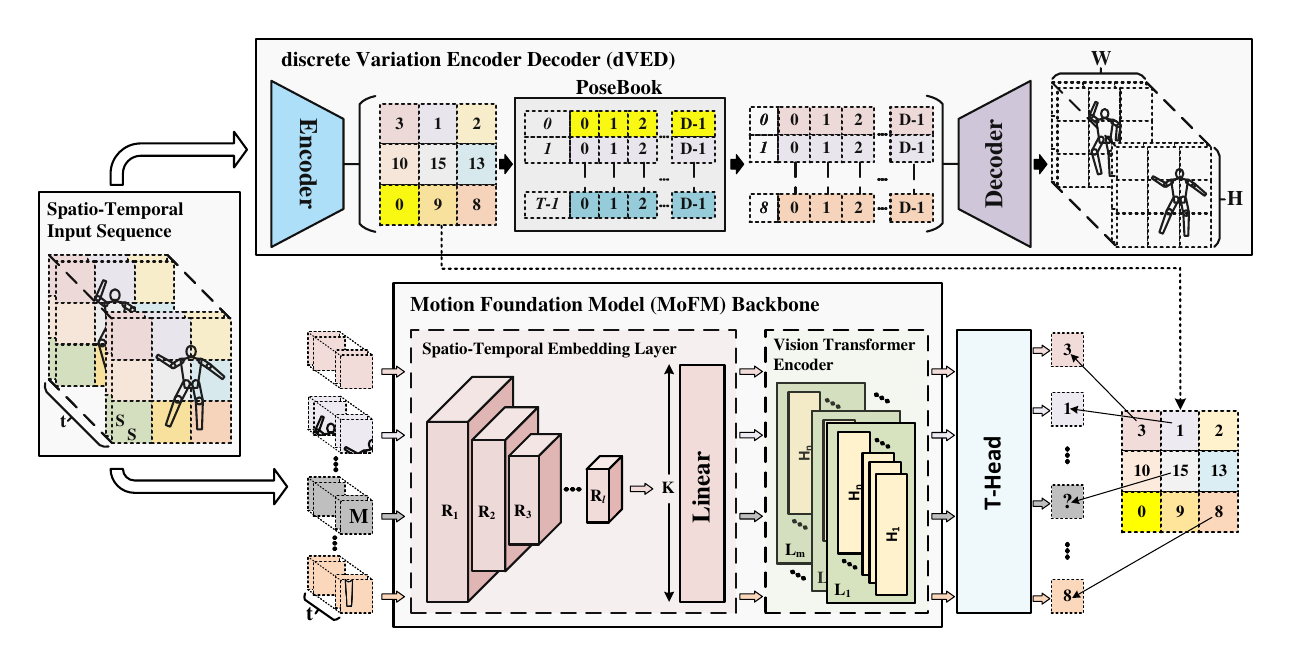}
  \caption{Overview of Motion Foundation Model (MoFM). Poses are converted into heatmap representations using a Gaussian function (\cref{eq:guss_factor}), producing a series of thermal cubes. Before pre-training, we train the custom dVED model for reconstruction. This involves tokenizing a series of heatmap cubes in both spatial and temporal dimensions according to a learned vocabulary. After cubing, tokens are masked keypoint-wise with a special mask embedding [\texttt{M}]. The resulting \(\{C_i^{m}\}_{i=0}^{K-1}\) masked cubes are then fed into a vision transformer encoder. The backbone predicts the visual tokens of the discretized image based on \(\{z_i\}_{i=0}^{K-1}\) generated by dVED.}
  \label{fig:self_trans}
\end{figure*}
\subsection{dVED for Motion Discretization}
In this paper, we present a custom dVED model with distinct encoder and decoder structures. The encoder utilizes 3D ResBlocks and 3D convolution blocks for dimensionality reduction, while the decoder employs 2D transposed convolutions and 2D ResBlocks to reconstruct inputs from the MotionBook back to \(U_R\) as illustrated in \cref{fig:dved}. 


The encoder, \({q}_{\theta}(z|U)\), is trained to map thermal cubes to a discrete latent space, $z$, while the decoder reconstructs the condensed heatmap, $U_{R}$, by applying \({p}_{\phi}(U_{R}|z)\). Therefore, The objective function can be formalized as follows:
\begin{align}
\label{eq:dv_loss}
\begin{split}
& \mathcal{L} = -\mathbb{E}_{{q}_{\theta}(z|U)}[\log {p}_{\phi}(U_{R}|z)] \\
& + \beta \times D_{\mathrm{KL}}({q}_{\theta}(z|U) \parallel p(z))
\end{split}
\end{align}
where \( \log {p}_{\phi}(U_{R}|z) \) represents the likelihood of the reconstructed data given the latent variables, \( D_{\mathrm{KL}} \) is the KL divergence between the approximate posterior \( {q}_{\theta}(z|U) \) and the prior \( p(z) \), typically chosen to be uniform over the discrete categories, and $\beta$ is the coefficient for the KL divergence term.

The Gumbel-Softmax trick allows this objective to be optimized via gradient descent by approximating categorical sampling using the softmax function:
\begin{align}
y_i = \frac{\exp((\log(\pi_i) + g_i) / \tau)}{\sum_{j=0}^{T-1} \exp((\log(\pi_j) + g_j) / \tau)},
\end{align}
where \( \tau \) is the temperature that controls the softness of the distribution. We schedule \(\tau\) to be dynamically reduced during the training phase to improve performance. For our setup, we approximate the reconstruction term \(\mathbb{E}_{{q}_{\theta}(z|U)}[\log {p}_{\phi}(U_{R}|z)]\) with the smooth L1 loss function.

For better generalization of dVED, it is essential that the poses encompass a wide variety of actions covering the full range of motion. We chose the CMU Panoptic dataset, a comprehensive collection offering sequences of pose information from multiple perspectives. For each randomly selected video from the dataset, we divided it into \(F\) bins and randomly selected a frame from each bin. We then applied \cref{eq:guss_factor} to generate \(U\), and \(U_R\). The dVED model architecture features an encoder with two 3D Convolution and two 3D ResNet blocks, and a decoder with two 2D Convolution and two 2D ResNet blocks, with a hidden dimension of 256. The pose book also has a 256 hidden dimension and a vocabulary size of 8192.
Training was performed over 20 epochs with a batch size of 24, utilizing the AdamW optimizer and a WarmupCosineLR scheduler. The maximum and minimum learning rates were set to $3.0e-4$ and $1.0e-6$, respectively, with a single warmup epoch. \cref{fig:combined_frames} qualitatively compares the performance of dVED, shown in \cref{fig:dvae}, with the ground truth, shown in \cref{fig:gt}. A notable phenomenon observed in the reconstructed heatmap is the ghosting effect around moving joints, which clearly demonstrates how the thermal energy transitions from one point to another in the direction of movement.

\subsection{MoFM Backbone and Self-supervised Training}
The Spatio-Temporal Embedding Layer of the MoFM consists of $\textit{l}$ layers of 3D residual block ($R$) to map $C$ to \(\mathcal{H} | \mathcal{H} \in \mathbb{R}^{K \times H}\) where $H$ is the hidden size of the vision transfomer encoder. The transformer encoder used in our work is the base transformer model proposed in \cite{dosovitskiy2021an, VaswaniTransformer}. The architecture uses 12 layers and 12 attention heads, each with a size of 64. The hidden dimension, feed-forward network size, and maximum sequence length are set to 768, 3072, and 384, respectively.
 
After training dVED, we used its encoder as a tokenizer. As outlined in \cref{subsec:pose_token}, to enable self-supervised training of the transformer encoder, we split the input \(U\) into \(K\) cubes, \(\{C_i\}_{i=0}^{K-1}\), and utilized the dVED tokenizer to map these cubes to the latent space, \(\{z_i\}_{i=0}^{K-1}\). Consequently, we set the transformer's input sequence length to \(K\). 

We propose a targeted approach for motion-wise masking of spatio-temporal cubes, detailed in Algorithm \ref{algo:blocking}. Initially, we experimented with blockwise masking as suggested in \cite{bao2022beit}; however, while this reduced the loss error, it did not yield qualitatively accurate reconstructions. We attribute this to the random block placements, which often masked regions without any pose skeletons, leading to ineffective learning. To address this, we developed keypoint-wise masking. In this approach, we randomly select a joint \( j \) from the given set \( P \). For each selected joint \( j \), located at coordinates \(\langle x_{jf}, y_{jf} \rangle\) in frame \( f \), we identify the block containing the joint and mark it as a candidate for masking. This process is repeated until the number of masked blocks specified by \( num\_block\_to\_mask \) in Algorithm \ref{algo:blocking} is reached.

The transformer encoder is then trained to classify the received Thermal Cubes based on the output of the dVED encoder. \cref{fig:self_trans} illustrates the closed-loop training process, in which the dVED encoder functions as a tokenizer to support self-supervised training. Masked cubes, represented as [\texttt{M}], are shaded in gray and treated as learnable parameters. Training was conducted over 20 epochs with a batch size of 32, using the AdamW optimizer with a cosine learning rate scheduler and warmup. The learning rate was set between a maximum of \(1.5 \times 10^{-4}\) and a minimum of \(1.0 \times 10^{-8}\), with a single warmup epoch.

\begin{algorithm*}[h]
\caption{Generate Mask from Keypoints}
 \label{algo:blocking}
\SetKwInOut{Input}{Input}
\SetKwInOut{Output}{Output}

\Input{$KP$: array of shape $(F \times J, 2)$ containing coordinate of keypoints in $[x, y]$  \\
       $H$: Height of the original heatmap \\
       $W$: Width of the original heatmap \\
       $D_S$: Size of spatial resolution \\
       $min\_masked\_number$: Minimum number of points to be randomly selected for masking \\
       $max\_masked\_number$: Maximum number of points to be randomly selected for masking}
\Output{$final\_mask$: A single mask of shape $(total\_rows\_squares, total\_cols\_squares)$ with 1s where squares are masked}

$total\_rows\_squares \gets \left\lfloor \frac{H}{S} \right\rfloor + \text{(1 if } H \% D_S \text{ else 0)}$\;

$total\_cols\_squares \gets \left\lfloor \frac{W}{S} \right\rfloor + \text{(1 if } W \% D_S \text{ else 0)}$\;

$final\_mask \gets \text{zeros}((total\_rows\_squares, total\_cols\_squares))$\;

$total\_points \gets F \times J$

$\alpha \gets \text{max}(1, \text{min}(min\_masked\_number, total\_points))$\;

$\beta \gets \text{min}(max\_masked\_number, total\_points)$\;

$num\_block\_to\_mask \gets \text{RandomSample}(\{\alpha, \alpha+1, \ldots, \beta, \beta + 1\}, 1)$\;


$selected\_indices \gets \text{RandomSample}\left(\{0, 1, 2, \ldots, total\_points - 1\}, num\_block\_to\_mask\right)$

$selected\_points \gets KP[selected\_indices]$\;

\ForEach{$(x, y) \in selected\_points$}{
    $col \gets \left\lfloor \frac{x}{D_S} \right\rfloor$\;
    
    $row \gets \left\lfloor \frac{y}{D_S} \right\rfloor$\;
    
    \If{$0 \leq row < total\_rows\_squares$ \textbf{and} $0 \leq col < total\_cols\_squares$}{
        $final\_mask[row, col] \gets 1$\;
    }
}

\Return $final\_mask$\;
\end{algorithm*}


\section{Experimental Results}
\label{sec:results}

This section presents experimental results to demonstrate the capabilities of the pretrained MoFM when used as a backbone for various downstream tasks. While MoFM is applicable to a wide range of applications, we have selected DT1: action classification, DT2: one-shot action classification, DT3: self-supervised human anomaly detection, and DT4: supervised human anomaly detection to showcase its versatility. In all these tasks, we have only added a basic fully connected linear head on top of MoFM to form a simple baseline. 

All training used six NVIDIA A6000 GPUs. Other experimental setup and hyper parameters for each downstream task are fully explained in the supplementary material.


\subsection{DT1: Action Classification} \label{sec:dt1}
We explore the potential of MoFM’s pretrained human motion representations as a backbone for learning action semantics. For this, we utilize the NTU-RGB+D \cite{ntu60Paper} dataset, containing 57K videos across 60 action classes, following the Cross-Subject (X-Sub) and Cross-View (X-View) splits. We also leverage the extended NTU-RGB+D-120 \cite{ntu120paper} dataset, which includes 114K videos covering 120 action classes with Cross-Subject (X-Sub) and Cross-Setup (X-Set) splits. For both datasets, we extract 2D skeletons using HRNet \cite{hrnet} as described in \cite{PoseConv3D}. As shown in \cref{tab:ntu_60} and \cref{tab:ntu_120}, our methods achieve performance on par with or surpassing SOTA approaches, underscoring MoFM’s capability to generalize effectively to the downstream task of human action recognition.

\begin{table}[htbp]
    \centering
    \caption{Fine-tuning results for action classification as the downstream task on NTU RGB+D \cite{ntu60Paper} dataset (top-1\% accuracy). MoFM-FC represents MoFM with a basic fully connected head to serve as a simple baseline.}
    \label{tab:ntu_60}
    \begin{tabular}{lccc}
        \toprule
        \toprule
        \textbf{Method} & \textbf{Source} & \textbf{X-Sub} & \textbf{X-View} \\
        \midrule
        Vemulapalli et al. \cite{vemulapalli2014human} & CVPR & 50.1 & 52.8 \\
        Liu et al. \cite{liu2016spatio} & ECCV & 69.2 & 77.7 \\
        Soo et al. \cite{soo2017interpretable} & CVPR & 74.3 & 83.1 \\
        Zhang et al. \cite{zhang2017view} & ICCV & 79.2 & 87.7 \\
        Li et al. \cite{li2018independently} & CVPR & 81.8 & 88.0 \\
        Cho et al. \cite{cho2020self} & WACV & 87.2 & 92.7 \\
        Li et al. \cite{li2019actional} & CVPR & 86.8 & 94.2 \\
        Song et al. \cite{song2020richly} & TCSVT & 87.3 & 93.6 \\
        Shi et al. \cite{shi2019two} & CVPR & 88.5 & 95.1 \\
        Shi et al. \cite{shi2019skeleton} & CVPR & 89.2 & 95.5 \\
        Chen et al. \cite{chen2021multi} & AAAI & 91.5 & 96.6 \\
        Miao et al. \cite{miao2021central} & TCSVT & 90.9 & 96.6 \\
        Xia et al. \cite{xia2021laga} & TMM & 87.1 & 93.2 \\
        Jang et al. \cite{jang2024multi} & TCSVT & 92.7 & 96.9 \\
        Liu et al. \cite{9156556} & CVPR & 91.5 & 96.2 \\
        Li et al. \cite{9577939} & CVPR & 86.2 & 92.5 \\
        Shi et al. \cite{twStreamShi2019} & CVPR & 88.5 & 95.1 \\
        Duan et al. \cite{PoseConv3D} & CVPR & 93.1 & 95.7 \\
        Su et al. \cite{9711113} & ICCV & 89.7 & 96.3 \\
        Yan et al. \cite{SpatialYan2018} & AAAI & 81.5 & 88.3 \\
        Yang et al. \cite{yang2021skeleton} & ICCV & 88.0 & 94.9 \\
        Cheng et al. \cite{9157077} & CVPR & 90.7 & 96.5 \\
        \midrule
        \memt-FC (Ours) & - & 89.4 & 94.7 \\
        \bottomrule
        \bottomrule
    \end{tabular}
\end{table}

\begin{table}[htbp]
    \centering
    \caption{Fine-tuning results for action classification as the downstream task on NTU RGB+D 120 \cite{ntu120paper} dataset (top-1\% accuracy). MoFM-FC represents MoFM with a basic fully connected head to serve as a simple baseline.}
    \label{tab:ntu_120}
    \begin{tabular}{lccc}
        \toprule
        \toprule
        \textbf{Methods} & \textbf{Source} & \textbf{X-Sub} & \textbf{X-Set} \\
        \midrule
        Yan et al. \cite{SpatialYan2018} & AAAI & 70.7 & 73.2 \\
        Li et al. \cite{li2019actional} & CVPR & 77.7 & 78.9 \\
        Song et al. \cite{song2020richly} & TCSVT & 81.1 & 82.7 \\
        Shi et al. \cite{shi2019two} & CVPR & 82.9 & 84.9 \\
        Liu et al. \cite{9156556} & CVPR & 86.9 & 88.4 \\
        Chen et al. \cite{chen2021multi} & AAAI & 87.5 & 88.8 \\
        Miao et al. \cite{miao2021central} & TCSVT & 86.3 & 87.8 \\
        Wu et al. \cite{wu2021graph2net} & TCSVT & 86.0 & 87.6 \\
        Xia et al. \cite{xia2021laga} & TMM & 81.0 & 82.2 \\
        Jang et al. \cite{jang2024multi} & TCSVT & 88.9 & 90.6 \\
        Liu et al. \cite{8099874} & CVPR & 67.7 & 66.9 \\
        Ke et al. \cite{8306456} & TIP & 62.2 & 61.8 \\
        Shahroudy et al. \cite{ntu60Paper} & CVPR & 25.5 & 26.3 \\
        Liu et al. \cite{liu2016spatio} & ECCV & 55.7 & 57.9 \\
        Caetano et al. \cite{8919860} & SIBGRAPI & 67.9 & 62.8 \\
        Cheng et al. \cite{9157077} & CVPR & 80.9 & 83.2 \\
        \midrule
        \memt-FC (Ours) & - & 78.0 & 83.7 \\
        \bottomrule
        \bottomrule
    \end{tabular}
\end{table}

\subsection{DT2: One-shot Action Classification}\label{sec:dt2}
We further assess MoFM's capability in recognizing one-shot actions by adding a fully connected layer that embeds the given action onto a $\delta$-dimensional hypersphere, where $\delta = 2048$ for this experiment. Following NTU-RGB+D-120 \cite{ntu120paper} guidelines, we trained the model on an \textit{auxiliary set} of 100 labeled samples using supervised contrastive learning \cite{khosla2020supervised}. In each batch, \( m \) samples of the same class were randomly selected from the auxiliary set, and a supervised contrastive loss was applied to cluster samples of the same class while separating samples of different classes. Evaluation was conducted by calculating cosine similarity between the test set and 20 exemplars for each remaining class. \cref{tab:ntu_120_one} presents the accuracy comparison of MoFM with recent approaches, demonstrating MoFM's effectiveness in handling one-shot action classification, similar to the DT1 experiments.

\begin{table}[htbp]
       \centering
       \caption{Fine-tuning results for one-shot action classification as the downstream task on NTU RGB+D 120 \cite{ntu120paper} dataset. MoFM-FC represents MoFM with a basic fully connected head to serve as a simple baseline.}
       \label{tab:ntu_120_one}
           \begin{tabular}{lccc}
               \toprule
               \toprule
               \textbf{Methods} & \textbf{Source} & \textbf{Top-1\%} \\
               \hline
               Liu et al. \cite{liu2017global} & CVPR& 41.0\\
               Liu et al. \cite{liu2017global} &CVPR & 42.1\\
               Liu et al. \cite{liu2017skeleton} & TPAMI& 42.9\\
               Liu et al. \cite{liu2019ntu} &TPAMI & 45.3\\
               Sabater et al. \cite{sabater2021one} & CVPR& 46.5\\
               Memmesheimer et al. \cite{memmesheimer2021sl} &ICPR & 50.9\\
               Memmesheimer et al. \cite{memmesheimer2022skeleton} & WACV& 54.2\\
               Wang et al. \cite{wang2022uncertainty} & ECCV& 49.0\\
               Wang et al. \cite{wang2022temporal} & ACCV& 57.0\\
               Zhu et al. \cite{zhu2023adaptive} & WACV& 57.6\\
               \hline
               \memt-FC (Ours) & -& 56.6\\
               \bottomrule
               \bottomrule
           \end{tabular}
\end{table}





\subsection{DT-3: Self-supervised Human Anomaly Detection}
\label{sec:sht}
In self-supervised human video anomaly detection, the objective is to enable models to discern patterns of typical human behavior and identify deviations as anomalous \cite{10.1145/3645101}. Self-supervised methods typically formulate proxy tasks, such as reconstruction \cite{morais2019learning} or future prediction \cite{morais2019learning, rodrigues2020multi, zeng2021hierarchical}, training the model exclusively on normal data. This encourages the model to internalize the underlying dynamics of normal human motion. During inference, the model's ability to perform these tasks serves as an indicator of abnormality detection.

\begin{table}[htbp]
    \centering
    \caption{Fine-tuning results for self-supervised human anomaly detection as the downstream task on ShanghaiTech Campus (SHT) \cite{liu2018future} and HR-ShanghaiTech (HR-SHT) \cite{morais2019learning} datasets (AUC-ROC\%). MoFM-FC represents MoFM with a basic fully connected head to serve as a simple baseline.}
    \label{tab:sht}
        \begin{tabular}{lccc}
            \toprule
            \toprule
            \textbf{Methods} & \textbf{Source} & \textbf{SHT} & \textbf{HR-SHT} \\
            \hline
            Wang et al. \cite{wang2023memory} & PR & 71.30 & - \\
            Zaheer et al. \cite{zaheer2022generative} & CVPR & 78-93 & - \\
            Luo et al. \cite{Luo_2017_ICCV} & ICCV & 68.00 & - \\
            Abati et al. \cite{abati2019latent} & CVPR & 72.50 & - \\
            Hasan et al. \cite{hasan2016learning} & CVPR & 70.40 & 69.80 \\
            Liu et al. \cite{Liu_2018_CVPR} & CVPR & 72.80 & 72.70 \\
            Morais et al. \cite{morais2019learning} & CVPR & 73.40 & 75.40 \\
            Markovitz et al. \cite{markovitz2020graph} & CVPR & 75.50 & - \\
            Jain et al. \cite{jain2021posecvae} & ICPR & 74.90 & 75.70 \\
            Chen et al. \cite{chen2023multiscale} & JVCIR & 75.90 & - \\
            Rodrigues et al. \cite{rodrigues2020multi} & WACV & 76.03 & 77.04 \\
            Huang et al. \cite{huang2022hierarchical} &CVPR  & 82.90 & 86.97 \\
            Noghre et al. \cite{Noghre_2024_WACV} & WACV & 80.67 & 81.77 \\
            \hline
            \memt-FC (Ours) & - & 76.56 & 77.26 \\
            \bottomrule
            \bottomrule
        \end{tabular}
\end{table}

Inspired by \cite{jigsaw}, we define a self-supervised jigsaw puzzle task at the token level, rather than at the pixel level as in \cite{jigsaw}. A simple fully connected layer with 12 neurons is added as the task head. Tokens are grouped into 12 puzzle pieces, which are shuffled during training, and the network is fine-tuned with multi-label supervision using cross-entropy loss to predict the correct positions of the pieces. During inference, no shuffling occurs; instead, the model's predictions are used to compute a normality score. For abnormal behavior, prediction accuracy declines, resulting in a non-diagonal probability matrix. The lowest probability along the diagonal is taken as the normality score. If multiple individuals are present in the scene, the frame’s overall normality score is determined by the minimum of their individual scores.

To evaluate our model, we use the widely adopted ShanghaiTech Campus (SHT) dataset \cite{liu2018future}. Following previous SOTA methods \citep{markovitz2020graph, hirschorn2023normalizing, yu2023regularity}, we use AlphaPose \citep{li2019crowdpose} for pose extraction and tracking to ensure fair comparison. We also report results on the Human-Related SHT subset \cite{morais2019learning}, which focuses exclusively on human activities. We report results in terms of frame level Area Under the Receiver Operating Characteristic Curve (AUC-ROC). 

As shown in \cref{tab:sht}, MoFM-FC achieves an AUC-ROC of 76.56\% on SHT \cite{liu2018future} and 77.26\% on HR-SHT \cite{morais2019learning}, demonstrating performance comparable to other SOTA models. These results underscore the adaptability and versatility of the MoFM as a backbone, demonstrating that with the addition of a simple linear head, it can be effectively utilized for self-supervised anomaly detection.

\subsection{DT4: Supervised Human Anomaly Detection}
\label{sec:sup_anomaly}
In supervised human anomaly detection, the task is framed as a binary classification problem. Similar to previous downstream tasks, we add a fully connected layer as the head. During inference, the probability assigned to the normal class serves as the normality score. In line with the self-supervised approach, if multiple individuals are present, the frame's overall normality score is determined by the minimum of their individual scores.

\begin{table}[htbp]
    \centering
    \caption{Fine-tuning results for supervised human anomaly detection as the downstream task on the UBnormal \cite{Acsintoae_CVPR_2022} dataset (AUC\%). MoFM-FC represents MoFM with a basic fully connected head to serve as a simple baseline.}
    \label{tab:ub}
        \begin{tabular}{lcc}
            \toprule
            \toprule
            \textbf{Methods} & \textbf{Source} & \textbf{AUC-ROC\%} \\
            \hline
            Georgescu et al. \cite{georgescu2021background} & TPAMI & 61.3 \\
            Bertasius et al. \cite{bertasius2021space} & ICML & 68.5 \\
            Yan et al. \cite{yan2018spatial} & AAAI & 78.1 \\
            Shi et al. \cite{shi2019two} & CVPR & 74.1 \\
            Cheng et al. \cite{cheng2020skeleton} & CVPR & 64.6 \\
            Liu et al. \cite{liu2020disentangling} & CVPR & 77.8 \\
            \hline
            \memt-FC (Ours) & - & 69.55 \\
            \bottomrule
            \bottomrule
        \end{tabular}
\end{table}

For evaluation, we utilize the UBnormal dataset \cite{Acsintoae_CVPR_2022} and measure the model's performance using frame-level AUC-ROC, consistent with prior SOTA approaches \cite{georgescu2021background, bertasius2021space, yan2018spatial, shi2019two, cheng2020skeleton, liu2020disentangling, hirschorn2023normalizing}. To address data imbalance in the training set, we apply data augmentation techniques, including random scaling, jitter noise, and horizontal flipping, to increase the representation of abnormal samples. Additionally, we use a weighted cross-entropy loss, giving 2$\times$ greater weight to abnormal samples to further mitigate class imbalance.

\cref{tab:ub} demonstrates the effectiveness of MoFM-FC compared to other SOTA models. Our model achieves an AUC-ROC of 69.55\%, showing performance comparable to other models and highlighting the capability of the MoFM backbone for supervised anomaly detection.

\section{Conclusion}
\label{sec:conslusion}

In this work we presented MoFM, a large-scale motion foundation model that understands the semantics of human motions. By utilizing Thermal Cubes and a discrete Variational Encoder-Decoder (dVED) model, we mapped spatio-temporal heatmaps into a discrete latent space, creating the comprehensive MotionBook dictionary. This approach enables task-agnostic self-supervised training, allowing MoFM to learn rich representations of human motion. Using MoFM as a foundational backbone, augmented with a simple fully connected linear layer to create a baseline, we have demonstrated its capabilities in handling downstream tasks including action classification, one-shot action classification, self-supervised human anomaly detection, and supervised human anomaly detection. We believe that MoFM will serve as a valuable foundation for future research endeavors in human motion understanding, enabling the development of more advanced models and applications in the field.

{
    \small
    \bibliographystyle{ieeenat_fullname}
    \bibliography{main}

\begin{thebibliography}{116}
\providecommand{\natexlab}[1]{#1}
\providecommand{\url}[1]{\texttt{#1}}
\expandafter\ifx\csname urlstyle\endcsname\relax
  \providecommand{\doi}[1]{doi: #1}\else
  \providecommand{\doi}{doi: \begingroup \urlstyle{rm}\Url}\fi

\bibitem[Abati et~al.(2019)Abati, Porrello, Calderara, and Cucchiara]{abati2019latent}
Davide Abati, Angelo Porrello, Simone Calderara, and Rita Cucchiara.
\newblock Latent space autoregression for novelty detection.
\newblock In \emph{Proceedings of the IEEE/CVF conference on computer vision and pattern recognition}, pages 481--490, 2019.

\bibitem[Acsintoae et~al.(2022)Acsintoae, Florescu, Georgescu, Mare, Sumedrea, Ionescu, Khan, and Shah]{Acsintoae_CVPR_2022}
Andra Acsintoae, Andrei Florescu, Mariana{-}Iuliana Georgescu, Tudor Mare, Paul Sumedrea, Radu~Tudor Ionescu, Fahad~Shahbaz Khan, and Mubarak Shah.
\newblock Ubnormal: New benchmark for supervised open-set video anomaly detection.
\newblock In \emph{Proceedings of the IEEE/CVF Conference on Computer Vision and Pattern Recognition (CVPR)}, 2022.

\bibitem[Ahn et~al.(2023)Ahn, Kim, Hong, and Chul~Ko]{10030830}
Dasom Ahn, Sangwon Kim, Hyunsu Hong, and Byoung Chul~Ko.
\newblock Star-transformer: A spatio-temporal cross attention transformer for human action recognition.
\newblock In \emph{2023 IEEE/CVF Winter Conference on Applications of Computer Vision (WACV)}, pages 3319--3328, 2023.

\bibitem[Alinezhad~Noghre et~al.(2023)Alinezhad~Noghre, Danesh~Pazho, Katariya, and Tabkhi]{alinezhad2023understanding}
Ghazal Alinezhad~Noghre, Armin Danesh~Pazho, Vinit Katariya, and Hamed Tabkhi.
\newblock Understanding the challenges and opportunities of pose-based anomaly detection.
\newblock In \emph{Proceedings of the 8th international Workshop on Sensor-Based Activity Recognition and Artificial Intelligence}, pages 1--9, 2023.

\bibitem[Ardabili et~al.(2023)Ardabili, Pazho, Noghre, Neff, Bhaskararayuni, Ravindran, Reid, and Tabkhi]{ardabili2023understanding}
Babak~Rahimi Ardabili, Armin~Danesh Pazho, Ghazal~Alinezhad Noghre, Christopher Neff, Sai~Datta Bhaskararayuni, Arun Ravindran, Shannon Reid, and Hamed Tabkhi.
\newblock Understanding policy and technical aspects of ai-enabled smart video surveillance to address public safety.
\newblock \emph{Computational Urban Science}, 3\penalty0 (1):\penalty0 21, 2023.

\bibitem[Bao et~al.(2021)Bao, Dong, and Wei]{bao2021beit}
Hangbo Bao, Li Dong, and Furu Wei.
\newblock {BEiT}: {BERT} pre-training of image transformers.
\newblock In \emph{International Conference on Learning Representations}, 2021.

\bibitem[Bao et~al.(2022)Bao, Dong, Piao, and Wei]{bao2022beit}
Hangbo Bao, Li Dong, Songhao Piao, and Furu Wei.
\newblock {BE}it: {BERT} pre-training of image transformers.
\newblock In \emph{International Conference on Learning Representations}, 2022.

\bibitem[Bertasius et~al.(2021)Bertasius, Wang, and Torresani]{bertasius2021space}
Gedas Bertasius, Heng Wang, and Lorenzo Torresani.
\newblock Is space-time attention all you need for video understanding?
\newblock In \emph{ICML}, page~4, 2021.

\bibitem[Biswas et~al.(2020)Biswas, Ghosh, and Ghosh]{biswas2020dvae}
Biswajit Biswas, Swarup~Kr Ghosh, and Anupam Ghosh.
\newblock Dvae: deep variational auto-encoders for denoising retinal fundus image.
\newblock \emph{Hybrid machine intelligence for medical image analysis}, pages 257--273, 2020.

\bibitem[Brown(2020)]{brown2020language}
Tom~B Brown.
\newblock Language models are few-shot learners.
\newblock \emph{arXiv preprint arXiv:2005.14165}, 2020.

\bibitem[Caetano et~al.(2019)Caetano, Bremond, and Schwartz]{8919860}
C. Caetano, F. Bremond, and W. Schwartz.
\newblock Skeleton image representation for 3d action recognition based on tree structure and reference joints.
\newblock In \emph{2019 32nd SIBGRAPI Conference on Graphics, Patterns and Images (SIBGRAPI)}, pages 16--23, Los Alamitos, CA, USA, 2019. IEEE Computer Society.

\bibitem[Chen et~al.(2022)]{chen2022internvl}
K. Chen et~al.
\newblock Intern{VL}: Vision-language pretraining with large-scale data.
\newblock \emph{arXiv preprint arXiv:2201.00000}, 2022.

\bibitem[Chen et~al.(2023)Chen, Kan, Zhang, Cen, Zhang, and Zhang]{chen2023multiscale}
Xiaoyu Chen, Shichao Kan, Fanghui Zhang, Yigang Cen, Linna Zhang, and Damin Zhang.
\newblock Multiscale spatial temporal attention graph convolution network for skeleton-based anomaly behavior detection.
\newblock \emph{Journal of Visual Communication and Image Representation}, 90:\penalty0 103707, 2023.

\bibitem[Chen et~al.(2021)Chen, Li, Yang, Li, and Liu]{chen2021multi}
Zhan Chen, Sicheng Li, Bing Yang, Qinghan Li, and Hong Liu.
\newblock Multi-scale spatial temporal graph convolutional network for skeleton-based action recognition.
\newblock In \emph{Proceedings of the AAAI conference on artificial intelligence}, pages 1113--1122, 2021.

\bibitem[Chen et~al.(2024)Chen, Wu, Wang, Su, Chen, Xing, Zhong, Zhang, Zhu, Lu, et~al.]{chen2024internvl}
Zhe Chen, Jiannan Wu, Wenhai Wang, Weijie Su, Guo Chen, Sen Xing, Muyan Zhong, Qinglong Zhang, Xizhou Zhu, Lewei Lu, et~al.
\newblock Internvl: Scaling up vision foundation models and aligning for generic visual-linguistic tasks.
\newblock In \emph{Proceedings of the IEEE/CVF Conference on Computer Vision and Pattern Recognition}, pages 24185--24198, 2024.

\bibitem[Cheng et~al.(2020{\natexlab{a}})Cheng, Xiao, Wang, Shi, Huang, and Zhang]{cheng2020higherhrnet}
Bowen Cheng, Bin Xiao, Jingdong Wang, Honghui Shi, Thomas~S Huang, and Lei Zhang.
\newblock Higherhrnet: Scale-aware representation learning for bottom-up human pose estimation.
\newblock In \emph{Proceedings of the IEEE/CVF conference on computer vision and pattern recognition}, pages 5386--5395, 2020{\natexlab{a}}.

\bibitem[Cheng et~al.(2020{\natexlab{b}})Cheng, Zhang, He, Chen, Cheng, and Lu]{9157077}
Ke Cheng, Yifan Zhang, Xiangyu He, Weihan Chen, Jian Cheng, and Hanqing Lu.
\newblock Skeleton-based action recognition with shift graph convolutional network.
\newblock In \emph{2020 IEEE/CVF Conference on Computer Vision and Pattern Recognition (CVPR)}, pages 180--189, 2020{\natexlab{b}}.

\bibitem[Cheng et~al.(2020{\natexlab{c}})Cheng, Zhang, He, Chen, Cheng, and Lu]{cheng2020skeleton}
Ke Cheng, Yifan Zhang, Xiangyu He, Weihan Chen, Jian Cheng, and Hanqing Lu.
\newblock Skeleton-based action recognition with shift graph convolutional network.
\newblock In \emph{Proceedings of the IEEE/CVF conference on computer vision and pattern recognition}, pages 183--192, 2020{\natexlab{c}}.

\bibitem[Cho et~al.(2020)Cho, Maqbool, Liu, and Foroosh]{cho2020self}
Sangwoo Cho, Muhammad Maqbool, Fei Liu, and Hassan Foroosh.
\newblock Self-attention network for skeleton-based human action recognition.
\newblock In \emph{Proceedings of the IEEE/CVF winter conference on applications of computer vision}, pages 635--644, 2020.

\bibitem[Deng et~al.(2024)Deng, Zhang, He, Chen, Shi, Xu, Fu, Zhang, Wang, Zhou, et~al.]{deng2024k2}
Cheng Deng, Tianhang Zhang, Zhongmou He, Qiyuan Chen, Yuanyuan Shi, Yi Xu, Luoyi Fu, Weinan Zhang, Xinbing Wang, Chenghu Zhou, et~al.
\newblock K2: A foundation language model for geoscience knowledge understanding and utilization.
\newblock In \emph{Proceedings of the 17th ACM International Conference on Web Search and Data Mining}, pages 161--170, 2024.

\bibitem[Devlin et~al.(2019)Devlin, Chang, Lee, and Toutanova]{Devlin2019BERTPO}
Jacob Devlin, Ming-Wei Chang, Kenton Lee, and Kristina Toutanova.
\newblock Bert: Pre-training of deep bidirectional transformers for language understanding.
\newblock In \emph{North American Chapter of the Association for Computational Linguistics}, 2019.

\bibitem[Dosovitskiy et~al.(2021)Dosovitskiy, Beyer, Kolesnikov, Weissenborn, Zhai, Unterthiner, Dehghani, Minderer, Heigold, Gelly, Uszkoreit, and Houlsby]{dosovitskiy2021an}
Alexey Dosovitskiy, Lucas Beyer, Alexander Kolesnikov, Dirk Weissenborn, Xiaohua Zhai, Thomas Unterthiner, Mostafa Dehghani, Matthias Minderer, Georg Heigold, Sylvain Gelly, Jakob Uszkoreit, and Neil Houlsby.
\newblock An image is worth 16x16 words: Transformers for image recognition at scale.
\newblock In \emph{International Conference on Learning Representations}, 2021.

\bibitem[Duan et~al.(2022)Duan, Zhao, Chen, Lin, and Dai]{PoseConv3D}
Haodong Duan, Yue Zhao, Kai Chen, Dahua Lin, and Bo Dai.
\newblock Revisiting skeleton-based action recognition.
\newblock In \emph{2022 IEEE/CVF Conference on Computer Vision and Pattern Recognition (CVPR)}, pages 2959--2968, 2022.

\bibitem[Eswaran and Khang(2024)]{eswaran2024artificial}
Ushaa Eswaran and Alex Khang.
\newblock Artificial intelligence (ai)-aided computer vision (cv) in healthcare system.
\newblock In \emph{Computer Vision and AI-Integrated IoT Technologies in the Medical Ecosystem}, pages 125--137. CRC Press, 2024.

\bibitem[Fang et~al.(2022)Fang, Li, Tang, Xu, Zhu, Xiu, Li, and Lu]{fang2022alphapose}
Hao-Shu Fang, Jiefeng Li, Hongyang Tang, Chao Xu, Haoyi Zhu, Yuliang Xiu, Yong-Lu Li, and Cewu Lu.
\newblock Alphapose: Whole-body regional multi-person pose estimation and tracking in real-time.
\newblock \emph{IEEE Transactions on Pattern Analysis and Machine Intelligence}, 45\penalty0 (6):\penalty0 7157--7173, 2022.

\bibitem[Feng et~al.(2023)Feng, Gao, Tse, Ma, and Chang]{feng2023diffpose}
Runyang Feng, Yixing Gao, Tze Ho~Elden Tse, Xueqing Ma, and Hyung~Jin Chang.
\newblock Diffpose: Spatiotemporal diffusion model for video-based human pose estimation.
\newblock In \emph{Proceedings of the IEEE/CVF International Conference on Computer Vision}, pages 14861--14872, 2023.

\bibitem[Geng et~al.(2023)Geng, Lu, Hu, Liu, and Lyu]{geng2023focusing}
Pei Geng, Xuequan Lu, Chunyu Hu, Hong Liu, and Lei Lyu.
\newblock Focusing fine-grained action by self-attention-enhanced graph neural networks with contrastive learning.
\newblock \emph{IEEE Transactions on Circuits and Systems for Video Technology}, 33\penalty0 (9):\penalty0 4754--4768, 2023.

\bibitem[Georgescu et~al.(2021)Georgescu, Ionescu, Khan, Popescu, and Shah]{georgescu2021background}
Mariana~Iuliana Georgescu, Radu~Tudor Ionescu, Fahad~Shahbaz Khan, Marius Popescu, and Mubarak Shah.
\newblock A background-agnostic framework with adversarial training for abnormal event detection in video.
\newblock \emph{IEEE transactions on pattern analysis and machine intelligence}, 44\penalty0 (9):\penalty0 4505--4523, 2021.

\bibitem[Hasan et~al.(2016)Hasan, Choi, Neumann, Roy-Chowdhury, and Davis]{hasan2016learning}
Mahmudul Hasan, Jonghyun Choi, Jan Neumann, Amit~K Roy-Chowdhury, and Larry~S Davis.
\newblock Learning temporal regularity in video sequences.
\newblock In \emph{Proceedings of the IEEE conference on computer vision and pattern recognition}, pages 733--742, 2016.

\bibitem[Hirschorn and Avidan(2023)]{hirschorn2023normalizing}
Or Hirschorn and Shai Avidan.
\newblock Normalizing flows for human pose anomaly detection.
\newblock In \emph{Proceedings of the IEEE/CVF International Conference on Computer Vision}, pages 13545--13554, 2023.

\bibitem[Huang et~al.(2022)Huang, Liu, Zhang, Liu, Wen, Xu, and Wang]{huang2022hierarchical}
Chao Huang, Yabo Liu, Zheng Zhang, Chengliang Liu, Jie Wen, Yong Xu, and Yaowei Wang.
\newblock Hierarchical graph embedded pose regularity learning via spatio-temporal transformer for abnormal behavior detection.
\newblock In \emph{Proceedings of the 30th ACM international conference on multimedia}, pages 307--315, 2022.

\bibitem[Jain et~al.(2021)Jain, Sharma, Velmurugan, and Banerjee]{jain2021posecvae}
Yashswi Jain, Ashvini~Kumar Sharma, Rajbabu Velmurugan, and Biplab Banerjee.
\newblock Posecvae: Anomalous human activity detection.
\newblock In \emph{2020 25th International Conference on Pattern Recognition (ICPR)}, pages 2927--2934. IEEE, 2021.

\bibitem[Jang et~al.(2017)Jang, Gu, and Poole]{jang2017categorical}
Eric Jang, Shixiang Gu, and Ben Poole.
\newblock Categorical reparameterization with gumbel-softmax.
\newblock In \emph{International Conference on Learning Representations}, 2017.

\bibitem[Jang et~al.(2024)Jang, Lee, Kim, Lee, Woo, and Lee]{jang2024multi}
Sungjun Jang, Heansung Lee, Woo~Jin Kim, Jungho Lee, Sungmin Woo, and Sangyoun Lee.
\newblock Multi-scale structural graph convolutional network for skeleton-based action recognition.
\newblock \emph{IEEE Transactions on Circuits and Systems for Video Technology}, 2024.

\bibitem[Ke et~al.(2018)Ke, Bennamoun, An, Sohel, and Boussaid]{8306456}
Qiuhong Ke, Mohammed Bennamoun, Senjian An, Ferdous Sohel, and Farid Boussaid.
\newblock Learning clip representations for skeleton-based 3d action recognition.
\newblock \emph{IEEE Transactions on Image Processing}, 27\penalty0 (6):\penalty0 2842--2855, 2018.

\bibitem[Kenton and Toutanova(2019)]{kenton2019bert}
Jacob Devlin Ming-Wei~Chang Kenton and Lee~Kristina Toutanova.
\newblock Bert: Pre-training of deep bidirectional transformers for language understanding.
\newblock In \emph{Proceedings of naacL-HLT}, page~2. Minneapolis, Minnesota, 2019.

\bibitem[Khang et~al.(2024)Khang, Abdullayev, Litvinova, Chumachenko, Alyar, and Anh]{khang2024application}
Alex Khang, Vugar Abdullayev, Eugenia Litvinova, Svetlana Chumachenko, Abuzarova~Vusala Alyar, and PTN Anh.
\newblock Application of computer vision (cv) in the healthcare ecosystem.
\newblock In \emph{Computer Vision and AI-Integrated IoT Technologies in the Medical Ecosystem}, pages 1--16. CRC Press, 2024.

\bibitem[Khosla et~al.(2020)Khosla, Teterwak, Wang, Sarna, Tian, Isola, Maschinot, Liu, and Krishnan]{khosla2020supervised}
Prannay Khosla, Piotr Teterwak, Chen Wang, Aaron Sarna, Yonglong Tian, Phillip Isola, Aaron Maschinot, Ce Liu, and Dilip Krishnan.
\newblock Supervised contrastive learning.
\newblock \emph{Advances in neural information processing systems}, 33:\penalty0 18661--18673, 2020.

\bibitem[Kirillov et~al.(2023)Kirillov, Mintun, Ravi, Mao, Rolland, Gustafson, Xiao, Whitehead, Berg, et~al.]{kirillov2023segment}
Alexander Kirillov, Eric Mintun, Nikhila Ravi, Hanzi Mao, Paul Rolland, Laura Gustafson, Tete Xiao, Spencer Whitehead, Alexander~C. Berg, et~al.
\newblock Segment anything.
\newblock \emph{arXiv preprint arXiv:2304.02643}, 2023.

\bibitem[Kocabas et~al.(2019)Kocabas, Karagoz, and Akbas]{kocabas2019self}
Muhammed Kocabas, Salih Karagoz, and Emre Akbas.
\newblock Self-supervised learning of 3d human pose using multi-view geometry.
\newblock In \emph{Proceedings of the IEEE/CVF conference on computer vision and pattern recognition}, pages 1077--1086, 2019.

\bibitem[Li et~al.(2019{\natexlab{a}})Li, Wang, Zhu, Mao, Fang, and Lu]{li2019crowdpose}
Jiefeng Li, Can Wang, Hao Zhu, Yihuan Mao, Hao-Shu Fang, and Cewu Lu.
\newblock Crowdpose: Efficient crowded scenes pose estimation and a new benchmark.
\newblock In \emph{Proceedings of the IEEE/CVF conference on computer vision and pattern recognition}, pages 10863--10872, 2019{\natexlab{a}}.

\bibitem[Li et~al.(2021)Li, Wang, Ni, Wang, Yang, and Zhang]{9577939}
Linguo Li, Minsi Wang, Bingbing Ni, Hang Wang, Jiancheng Yang, and Wenjun Zhang.
\newblock 3d human action representation learning via cross-view consistency pursuit.
\newblock In \emph{2021 IEEE/CVF Conference on Computer Vision and Pattern Recognition (CVPR)}, pages 4739--4748, 2021.

\bibitem[Li et~al.(2019{\natexlab{b}})Li, Chen, Chen, Zhang, Wang, and Tian]{li2019actional}
Maosen Li, Siheng Chen, Xu Chen, Ya Zhang, Yanfeng Wang, and Qi Tian.
\newblock Actional-structural graph convolutional networks for skeleton-based action recognition.
\newblock In \emph{Proceedings of the IEEE/CVF conference on computer vision and pattern recognition}, pages 3595--3603, 2019{\natexlab{b}}.

\bibitem[Li et~al.(2018)Li, Li, Cook, Zhu, and Gao]{li2018independently}
Shuai Li, Wanqing Li, Chris Cook, Ce Zhu, and Yanbo Gao.
\newblock Independently recurrent neural network (indrnn): Building a longer and deeper rnn.
\newblock In \emph{Proceedings of the IEEE conference on computer vision and pattern recognition}, pages 5457--5466, 2018.

\bibitem[Lin et~al.(2021)Lin, Wang, and Liu]{LinWang2021}
K. Lin, L. Wang, and Z. Liu.
\newblock Mesh graphormer.
\newblock In \emph{2021 IEEE/CVF International Conference on Computer Vision (ICCV)}, pages 12919--12928, Los Alamitos, CA, USA, 2021. IEEE Computer Society.

\bibitem[Liu et~al.(2016)Liu, Shahroudy, Xu, and Wang]{liu2016spatio}
Jun Liu, Amir Shahroudy, Dong Xu, and Gang Wang.
\newblock Spatio-temporal lstm with trust gates for 3d human action recognition.
\newblock In \emph{Computer Vision--ECCV 2016: 14th European Conference, Amsterdam, The Netherlands, October 11-14, 2016, Proceedings, Part III 14}, pages 816--833. Springer, 2016.

\bibitem[Liu et~al.(2017{\natexlab{a}})Liu, Shahroudy, Xu, Kot, and Wang]{liu2017skeleton}
Jun Liu, Amir Shahroudy, Dong Xu, Alex~C Kot, and Gang Wang.
\newblock Skeleton-based action recognition using spatio-temporal lstm network with trust gates.
\newblock \emph{IEEE transactions on pattern analysis and machine intelligence}, 40\penalty0 (12):\penalty0 3007--3021, 2017{\natexlab{a}}.

\bibitem[Liu et~al.(2017{\natexlab{b}})Liu, Wang, Hu, Duan, and Kot]{8099874}
Jun Liu, Gang Wang, Ping Hu, Ling-Yu Duan, and Alex~C. Kot.
\newblock Global context-aware attention lstm networks for 3d action recognition.
\newblock In \emph{2017 IEEE Conference on Computer Vision and Pattern Recognition (CVPR)}, pages 3671--3680, 2017{\natexlab{b}}.

\bibitem[Liu et~al.(2017{\natexlab{c}})Liu, Wang, Hu, Duan, and Kot]{liu2017global}
Jun Liu, Gang Wang, Ping Hu, Ling-Yu Duan, and Alex~C Kot.
\newblock Global context-aware attention lstm networks for 3d action recognition.
\newblock In \emph{Proceedings of the IEEE conference on computer vision and pattern recognition}, pages 1647--1656, 2017{\natexlab{c}}.

\bibitem[Liu et~al.(2019)Liu, Shahroudy, Perez, Wang, Duan, and Kot]{liu2019ntu}
Jun Liu, Amir Shahroudy, Mauricio Perez, Gang Wang, Ling-Yu Duan, and Alex~C Kot.
\newblock Ntu rgb+ d 120: A large-scale benchmark for 3d human activity understanding.
\newblock \emph{IEEE transactions on pattern analysis and machine intelligence}, 42\penalty0 (10):\penalty0 2684--2701, 2019.

\bibitem[Liu et~al.(2020{\natexlab{a}})Liu, Shahroudy, Perez, Wang, Duan, and Kot]{ntu120paper}
Jun Liu, Amir Shahroudy, Mauricio Perez, Gang Wang, Ling-Yu Duan, and Alex~C. Kot.
\newblock Ntu rgb+d 120: A large-scale benchmark for 3d human activity understanding.
\newblock \emph{IEEE Trans. Pattern Anal. Mach. Intell.}, 42\penalty0 (10):\penalty0 2684–2701, 2020{\natexlab{a}}.

\bibitem[Liu et~al.(2018{\natexlab{a}})Liu, Luo, Lian, and Gao]{Liu_2018_CVPR}
Wen Liu, Weixin Luo, Dongze Lian, and Shenghua Gao.
\newblock Future frame prediction for anomaly detection – a new baseline.
\newblock In \emph{Proceedings of the IEEE Conference on Computer Vision and Pattern Recognition (CVPR)}, 2018{\natexlab{a}}.

\bibitem[Liu et~al.(2018{\natexlab{b}})Liu, Luo, Lian, and Gao]{liu2018future}
Wen Liu, Weixin Luo, Dongze Lian, and Shenghua Gao.
\newblock Future frame prediction for anomaly detection--a new baseline.
\newblock In \emph{Proceedings of the IEEE conference on computer vision and pattern recognition}, pages 6536--6545, 2018{\natexlab{b}}.

\bibitem[Liu et~al.(2024)Liu, Yang, Wang, Liu, Liu, Boukerche, Sun, and Song]{10.1145/3645101}
Yang Liu, Dingkang Yang, Yan Wang, Jing Liu, Jun Liu, Azzedine Boukerche, Peng Sun, and Liang Song.
\newblock Generalized video anomaly event detection: Systematic taxonomy and comparison of deep models.
\newblock 56\penalty0 (7), 2024.

\bibitem[Liu et~al.(2020{\natexlab{b}})Liu, Zhang, Chen, Wang, and Ouyang]{9156556}
Ziyu Liu, Hongwen Zhang, Zhenghao Chen, Zhiyong Wang, and Wanli Ouyang.
\newblock Disentangling and unifying graph convolutions for skeleton-based action recognition.
\newblock In \emph{2020 IEEE/CVF Conference on Computer Vision and Pattern Recognition (CVPR)}, pages 140--149, 2020{\natexlab{b}}.

\bibitem[Liu et~al.(2020{\natexlab{c}})Liu, Zhang, Chen, Wang, and Ouyang]{liu2020disentangling}
Ziyu Liu, Hongwen Zhang, Zhenghao Chen, Zhiyong Wang, and Wanli Ouyang.
\newblock Disentangling and unifying graph convolutions for skeleton-based action recognition.
\newblock In \emph{Proceedings of the IEEE/CVF conference on computer vision and pattern recognition}, pages 143--152, 2020{\natexlab{c}}.

\bibitem[Luo et~al.(2023)Luo, Kong, Zhang, Hu, Li, and Tang]{luo2023computer}
Kui Luo, Xuan Kong, Jie Zhang, Jiexuan Hu, Jinzhao Li, and Hao Tang.
\newblock Computer vision-based bridge inspection and monitoring: A review.
\newblock \emph{Sensors}, 23\penalty0 (18):\penalty0 7863, 2023.

\bibitem[Luo et~al.(2017)Luo, Liu, and Gao]{Luo_2017_ICCV}
Weixin Luo, Wen Liu, and Shenghua Gao.
\newblock A revisit of sparse coding based anomaly detection in stacked rnn framework.
\newblock In \emph{Proceedings of the IEEE International Conference on Computer Vision (ICCV)}, 2017.

\bibitem[Luo et~al.(2021)Luo, Wang, Huang, Wang, Tan, and Zhou]{luo2021rethinking}
Zhengxiong Luo, Zhicheng Wang, Yan Huang, Liang Wang, Tieniu Tan, and Erjin Zhou.
\newblock Rethinking the heatmap regression for bottom-up human pose estimation.
\newblock In \emph{Proceedings of the IEEE/CVF conference on computer vision and pattern recognition}, pages 13264--13273, 2021.

\bibitem[Ma et~al.(2023)Ma, Hong, Gul, Gandhi, Gao, and Krishna]{ma2023crepe}
Zixian Ma, Jerry Hong, Mustafa~Omer Gul, Mona Gandhi, Irena Gao, and Ranjay Krishna.
\newblock Crepe: Can vision-language foundation models reason compositionally?
\newblock In \emph{Proceedings of the IEEE/CVF Conference on Computer Vision and Pattern Recognition}, pages 10910--10921, 2023.

\bibitem[Markovitz et~al.(2020)Markovitz, Sharir, Friedman, Zelnik-Manor, and Avidan]{markovitz2020graph}
Amir Markovitz, Gilad Sharir, Itamar Friedman, Lihi Zelnik-Manor, and Shai Avidan.
\newblock Graph embedded pose clustering for anomaly detection.
\newblock In \emph{Proceedings of the IEEE/CVF Conference on Computer Vision and Pattern Recognition}, pages 10539--10547, 2020.

\bibitem[Memmesheimer et~al.(2021)Memmesheimer, Theisen, and Paulus]{memmesheimer2021sl}
Raphael Memmesheimer, Nick Theisen, and Dietrich Paulus.
\newblock Sl-dml: Signal level deep metric learning for multimodal one-shot action recognition.
\newblock In \emph{2020 25th International conference on pattern recognition (ICPR)}, pages 4573--4580. IEEE, 2021.

\bibitem[Memmesheimer et~al.(2022)Memmesheimer, H{\"a}ring, Theisen, and Paulus]{memmesheimer2022skeleton}
Raphael Memmesheimer, Simon H{\"a}ring, Nick Theisen, and Dietrich Paulus.
\newblock Skeleton-dml: Deep metric learning for skeleton-based one-shot action recognition.
\newblock In \emph{Proceedings of the IEEE/CVF Winter Conference on Applications of Computer Vision}, pages 3702--3710, 2022.

\bibitem[Miao et~al.(2021)Miao, Hou, Gao, Xu, and Li]{miao2021central}
Shuangyan Miao, Yonghong Hou, Zhimin Gao, Mingliang Xu, and Wanqing Li.
\newblock A central difference graph convolutional operator for skeleton-based action recognition.
\newblock \emph{IEEE Transactions on Circuits and Systems for Video Technology}, 32\penalty0 (7):\penalty0 4893--4899, 2021.

\bibitem[Morais et~al.(2019)Morais, Le, Tran, Saha, Mansour, and Venkatesh]{morais2019learning}
Romero Morais, Vuong Le, Truyen Tran, Budhaditya Saha, Moussa Mansour, and Svetha Venkatesh.
\newblock Learning regularity in skeleton trajectories for anomaly detection in videos.
\newblock In \emph{Proceedings of the IEEE/CVF conference on computer vision and pattern recognition}, pages 11996--12004, 2019.

\bibitem[Myers et~al.(2024)Myers, Mohawesh, Chellaboina, Sathvik, Venkatesh, Ho, Henshaw, Alhawawreh, Berdik, and Jararweh]{myers2024foundation}
Devon Myers, Rami Mohawesh, Venkata~Ishwarya Chellaboina, Anantha~Lakshmi Sathvik, Praveen Venkatesh, Yi-Hui Ho, Hanna Henshaw, Muna Alhawawreh, David Berdik, and Yaser Jararweh.
\newblock Foundation and large language models: fundamentals, challenges, opportunities, and social impacts.
\newblock \emph{Cluster Computing}, 27\penalty0 (1):\penalty0 1--26, 2024.

\bibitem[Neff et~al.(2019)Neff, Mendieta, Mohan, Baharani, Rogers, and Tabkhi]{neff2019revamp}
Christopher Neff, Mat{\'\i}as Mendieta, Shrey Mohan, Mohammadreza Baharani, Samuel Rogers, and Hamed Tabkhi.
\newblock Revamp 2 t: real-time edge video analytics for multicamera privacy-aware pedestrian tracking.
\newblock \emph{IEEE Internet of Things Journal}, 7\penalty0 (4):\penalty0 2591--2602, 2019.

\bibitem[Noghre et~al.(2024{\natexlab{a}})Noghre, Pazho, and Tabkhi]{Noghre_2024_WACV}
Ghazal~Alinezhad Noghre, Armin~Danesh Pazho, and Hamed Tabkhi.
\newblock An exploratory study on human-centric video anomaly detection through variational autoencoders and trajectory prediction.
\newblock In \emph{Proceedings of the IEEE/CVF Winter Conference on Applications of Computer Vision (WACV) Workshops}, pages 995--1004, 2024{\natexlab{a}}.

\bibitem[Noghre et~al.(2024{\natexlab{b}})Noghre, Pazho, and Tabkhi]{noghre2024exploratory}
Ghazal~Alinezhad Noghre, Armin~Danesh Pazho, and Hamed Tabkhi.
\newblock An exploratory study on human-centric video anomaly detection through variational autoencoders and trajectory prediction.
\newblock In \emph{Proceedings of the IEEE/CVF Winter Conference on Applications of Computer Vision}, pages 995--1004, 2024{\natexlab{b}}.

\bibitem[Pavlakos et~al.(2018)Pavlakos, Zhou, and Daniilidis]{Pavlakos2018}
Georgios Pavlakos, Xiaowei Zhou, and Kostas Daniilidis.
\newblock Ordinal depth supervision for 3d human pose estimation.
\newblock In \emph{2018 IEEE/CVF Conference on Computer Vision and Pattern Recognition}, pages 7307--7316, 2018.

\bibitem[Pazho et~al.(2023)Pazho, Neff, Noghre, Ardabili, Yao, Baharani, and Tabkhi]{pazho2023ancilia}
Armin~Danesh Pazho, Christopher Neff, Ghazal~Alinezhad Noghre, Babak~Rahimi Ardabili, Shanle Yao, Mohammadreza Baharani, and Hamed Tabkhi.
\newblock Ancilia: Scalable intelligent video surveillance for the artificial intelligence of things.
\newblock \emph{IEEE Internet of Things Journal}, 10\penalty0 (17):\penalty0 14940--14951, 2023.

\bibitem[Ramesh et~al.(2021)Ramesh, Pavlov, Goh, Gray, Voss, Radford, Chen, and Sutskever]{RameshZSdVAE}
Aditya Ramesh, Mikhail Pavlov, Gabriel Goh, Scott Gray, Chelsea Voss, Alec Radford, Mark Chen, and Ilya Sutskever.
\newblock Zero-shot text-to-image generation.
\newblock In \emph{Proceedings of the 38th International Conference on Machine Learning}, pages 8821--8831. PMLR, 2021.

\bibitem[Rasheed et~al.(2023)Rasheed, Khattak, Maaz, Khan, and Khan]{rasheed2023fine}
Hanoona Rasheed, Muhammad~Uzair Khattak, Muhammad Maaz, Salman Khan, and Fahad~Shahbaz Khan.
\newblock Fine-tuned clip models are efficient video learners.
\newblock In \emph{Proceedings of the IEEE/CVF Conference on Computer Vision and Pattern Recognition}, pages 6545--6554, 2023.

\bibitem[Razavi et~al.(2019)Razavi, van~den Oord, and Vinyals]{VqVAE2}
Ali Razavi, A\"{a}ron van~den Oord, and Oriol Vinyals.
\newblock \emph{Generating diverse high-fidelity images with VQ-VAE-2}.
\newblock Curran Associates Inc., Red Hook, NY, USA, 2019.

\bibitem[Rodrigues et~al.(2020)Rodrigues, Bhargava, Velmurugan, and Chaudhuri]{rodrigues2020multi}
Royston Rodrigues, Neha Bhargava, Rajbabu Velmurugan, and Subhasis Chaudhuri.
\newblock Multi-timescale trajectory prediction for abnormal human activity detection.
\newblock In \emph{Proceedings of the IEEE/CVF winter conference on applications of computer vision}, pages 2626--2634, 2020.

\bibitem[Sabater et~al.(2021)Sabater, Santos, Santos-Victor, Bernardino, Montesano, and Murillo]{sabater2021one}
Alberto Sabater, Laura Santos, Jose Santos-Victor, Alexandre Bernardino, Luis Montesano, and Ana~C Murillo.
\newblock One-shot action recognition in challenging therapy scenarios.
\newblock In \emph{Proceedings of the IEEE/CVF conference on computer vision and pattern recognition}, pages 2777--2785, 2021.

\bibitem[S\'ar\'andi et~al.(2021)S\'ar\'andi, Linder, Arras, and Leibe]{sarandi2021metrabs}
Istv\'an S\'ar\'andi, Timm Linder, Kai~O. Arras, and Bastian Leibe.
\newblock {MeTRAbs:} metric-scale truncation-robust heatmaps for absolute 3{D} human pose estimation.
\newblock \emph{IEEE Transactions on Biometrics, Behavior, and Identity Science}, 3\penalty0 (1):\penalty0 16--30, 2021.

\bibitem[Shahroudy et~al.(2016)Shahroudy, Liu, Ng, and Wang]{ntu60Paper}
A. Shahroudy, J. Liu, T. Ng, and G. Wang.
\newblock Ntu rgb+d: A large scale dataset for 3d human activity analysis.
\newblock In \emph{2016 IEEE Conference on Computer Vision and Pattern Recognition (CVPR)}, pages 1010--1019, Los Alamitos, CA, USA, 2016. IEEE Computer Society.

\bibitem[Shi et~al.(2019{\natexlab{a}})Shi, Zhang, Cheng, and Lu]{shi2019skeleton}
Lei Shi, Yifan Zhang, Jian Cheng, and Hanqing Lu.
\newblock Skeleton-based action recognition with directed graph neural networks.
\newblock In \emph{Proceedings of the IEEE/CVF conference on computer vision and pattern recognition}, pages 7912--7921, 2019{\natexlab{a}}.

\bibitem[Shi et~al.(2019{\natexlab{b}})Shi, Zhang, Cheng, and Lu]{shi2019two}
Lei Shi, Yifan Zhang, Jian Cheng, and Hanqing Lu.
\newblock Two-stream adaptive graph convolutional networks for skeleton-based action recognition.
\newblock In \emph{Proceedings of the IEEE/CVF conference on computer vision and pattern recognition}, pages 12026--12035, 2019{\natexlab{b}}.

\bibitem[Shi et~al.(2019{\natexlab{c}})Shi, Zhang, Cheng, and Lu]{twStreamShi2019}
L. Shi, Y. Zhang, J. Cheng, and H. Lu.
\newblock Two-stream adaptive graph convolutional networks for skeleton-based action recognition.
\newblock In \emph{2019 IEEE/CVF Conference on Computer Vision and Pattern Recognition (CVPR)}, pages 12018--12027, Los Alamitos, CA, USA, 2019{\natexlab{c}}. IEEE Computer Society.

\bibitem[Song et~al.(2020)Song, Zhang, Shan, and Wang]{song2020richly}
Yi-Fan Song, Zhang Zhang, Caifeng Shan, and Liang Wang.
\newblock Richly activated graph convolutional network for robust skeleton-based action recognition.
\newblock \emph{IEEE Transactions on Circuits and Systems for Video Technology}, 31\penalty0 (5):\penalty0 1915--1925, 2020.

\bibitem[Soo~Kim and Reiter(2017)]{soo2017interpretable}
Tae Soo~Kim and Austin Reiter.
\newblock Interpretable 3d human action analysis with temporal convolutional networks.
\newblock In \emph{Proceedings of the IEEE conference on computer vision and pattern recognition workshops}, pages 20--28, 2017.

\bibitem[Su et~al.(2021)Su, Lin, and Wu]{9711113}
Yukun Su, Guosheng Lin, and Qingyao Wu.
\newblock Self-supervised 3d skeleton action representation learning with motion consistency and continuity.
\newblock In \emph{2021 IEEE/CVF International Conference on Computer Vision (ICCV)}, pages 13308--13318, 2021.

\bibitem[Sun et~al.(2019{\natexlab{a}})Sun, Xiao, Liu, and Wang]{hrnet}
Ke Sun, Bin Xiao, Dong Liu, and Jingdong Wang.
\newblock Deep high-resolution representation learning for human pose estimation.
\newblock In \emph{2019 IEEE/CVF Conference on Computer Vision and Pattern Recognition (CVPR)}, pages 5686--5696, 2019{\natexlab{a}}.

\bibitem[Sun et~al.(2019{\natexlab{b}})Sun, Xiao, Liu, and Wang]{sun2019deep}
Ke Sun, Bin Xiao, Dong Liu, and Jingdong Wang.
\newblock Deep high-resolution representation learning for human pose estimation.
\newblock In \emph{Proceedings of the IEEE/CVF conference on computer vision and pattern recognition}, pages 5693--5703, 2019{\natexlab{b}}.

\bibitem[Vahdat et~al.(2018{\natexlab{a}})Vahdat, Andriyash, and Macready]{vahdat2018dvae}
Arash Vahdat, Evgeny Andriyash, and William Macready.
\newblock Dvae\#: Discrete variational autoencoders with relaxed boltzmann priors.
\newblock \emph{Advances in Neural Information Processing Systems}, 31, 2018{\natexlab{a}}.

\bibitem[Vahdat et~al.(2018{\natexlab{b}})Vahdat, Macready, Bian, Khoshaman, and Andriyash]{vahdat2018dvae++}
Arash Vahdat, William Macready, Zhengbing Bian, Amir Khoshaman, and Evgeny Andriyash.
\newblock Dvae++: Discrete variational autoencoders with overlapping transformations.
\newblock In \emph{International conference on machine learning}, pages 5035--5044. PMLR, 2018{\natexlab{b}}.

\bibitem[van~den Oord et~al.(2017)van~den Oord, Vinyals, and Kavukcuoglu]{VqVAE}
Aaron van~den Oord, Oriol Vinyals, and Koray Kavukcuoglu.
\newblock Neural discrete representation learning.
\newblock In \emph{Proceedings of the 31st International Conference on Neural Information Processing Systems}, page 6309–6318, Red Hook, NY, USA, 2017. Curran Associates Inc.

\bibitem[Vaswani et~al.(2017{\natexlab{a}})Vaswani, Shazeer, Parmar, Uszkoreit, Jones, Gomez, Kaiser, and Polosukhin]{VaswaniTransformer}
Ashish Vaswani, Noam Shazeer, Niki Parmar, Jakob Uszkoreit, Llion Jones, Aidan~N. Gomez, \L{}ukasz Kaiser, and Illia Polosukhin.
\newblock Attention is all you need.
\newblock In \emph{Proceedings of the 31st International Conference on Neural Information Processing Systems}, page 6000–6010, Red Hook, NY, USA, 2017{\natexlab{a}}. Curran Associates Inc.

\bibitem[Vaswani et~al.(2017{\natexlab{b}})Vaswani, Shazeer, Parmar, Uszkoreit, Jones, Gomez, Kaiser, and Polosukhin]{vaswani2017attention}
Ashish Vaswani, Noam Shazeer, Niki Parmar, Jakob Uszkoreit, Llion Jones, Aidan~N. Gomez, {\L}ukasz Kaiser, and Illia Polosukhin.
\newblock Attention is all you need.
\newblock In \emph{Advances in Neural Information Processing Systems}, pages 5998--6008, 2017{\natexlab{b}}.

\bibitem[Vemulapalli et~al.(2014)Vemulapalli, Arrate, and Chellappa]{vemulapalli2014human}
Raviteja Vemulapalli, Felipe Arrate, and Rama Chellappa.
\newblock Human action recognition by representing 3d skeletons as points in a lie group.
\newblock In \emph{Proceedings of the IEEE conference on computer vision and pattern recognition}, pages 588--595, 2014.

\bibitem[Wang et~al.(2022)Wang, Wang, Qin, Zhang, Bao, and Huang]{jigsaw}
Guodong Wang, Yunhong Wang, Jie Qin, Dongming Zhang, Xiuguo Bao, and Di Huang.
\newblock Video anomaly detection by solving decoupled spatio-temporal jigsaw puzzles.
\newblock In \emph{Computer Vision -- ECCV 2022}, pages 494--511, Cham, 2022. Springer Nature Switzerland.

\bibitem[Wang et~al.(2024)Wang, Vasu, Faghri, Vemulapalli, Farajtabar, Mehta, Rastegari, Tuzel, and Pouransari]{wang2024sam}
Haoxiang Wang, Pavan Kumar~Anasosalu Vasu, Fartash Faghri, Raviteja Vemulapalli, Mehrdad Farajtabar, Sachin Mehta, Mohammad Rastegari, Oncel Tuzel, and Hadi Pouransari.
\newblock Sam-clip: Merging vision foundation models towards semantic and spatial understanding.
\newblock In \emph{Proceedings of the IEEE/CVF Conference on Computer Vision and Pattern Recognition}, pages 3635--3647, 2024.

\bibitem[Wang and Koniusz(2022{\natexlab{a}})]{wang2022temporal}
Lei Wang and Piotr Koniusz.
\newblock Temporal-viewpoint transportation plan for skeletal few-shot action recognition.
\newblock In \emph{Proceedings of the Asian Conference on Computer Vision}, pages 4176--4193, 2022{\natexlab{a}}.

\bibitem[Wang and Koniusz(2022{\natexlab{b}})]{wang2022uncertainty}
Lei Wang and Piotr Koniusz.
\newblock Uncertainty-dtw for time series and sequences.
\newblock In \emph{European Conference on Computer Vision}, pages 176--195. Springer, 2022{\natexlab{b}}.

\bibitem[Wang et~al.(2023{\natexlab{a}})Wang, Tian, Zhou, Shi, and Hua]{wang2023memory}
Le Wang, Junwen Tian, Sanping Zhou, Haoyue Shi, and Gang Hua.
\newblock Memory-augmented appearance-motion network for video anomaly detection.
\newblock \emph{Pattern Recognition}, 138:\penalty0 109335, 2023{\natexlab{a}}.

\bibitem[Wang et~al.(2023{\natexlab{b}})Wang, Dai, Chen, Huang, Li, Zhu, Hu, Lu, Lu, Li, et~al.]{wang2023internimage}
Wenhai Wang, Jifeng Dai, Zhe Chen, Zhenhang Huang, Zhiqi Li, Xizhou Zhu, Xiaowei Hu, Tong Lu, Lewei Lu, Hongsheng Li, et~al.
\newblock Internimage: Exploring large-scale vision foundation models with deformable convolutions.
\newblock In \emph{Proceedings of the IEEE/CVF conference on computer vision and pattern recognition}, pages 14408--14419, 2023{\natexlab{b}}.

\bibitem[Wehrbein et~al.(2021)Wehrbein, Rudolph, Rosenhahn, and Wandt]{Wehrbein2021}
Tom Wehrbein, Marco Rudolph, Bodo Rosenhahn, and Bastian Wandt.
\newblock Probabilistic monocular 3d human pose estimation with normalizing flows.
\newblock In \emph{2021 IEEE/CVF International Conference on Computer Vision (ICCV)}, pages 11179--11188, 2021.

\bibitem[Wu et~al.(2021)Wu, Wu, and Kittler]{wu2021graph2net}
Cong Wu, Xiao-Jun Wu, and Josef Kittler.
\newblock Graph2net: Perceptually-enriched graph learning for skeleton-based action recognition.
\newblock \emph{IEEE transactions on circuits and systems for video technology}, 32\penalty0 (4):\penalty0 2120--2132, 2021.

\bibitem[Xia et~al.(2021)Xia, Li, and Luo]{xia2021laga}
Rongjie Xia, Yanshan Li, and Wenhan Luo.
\newblock Laga-net: Local-and-global attention network for skeleton based action recognition.
\newblock \emph{IEEE Transactions on Multimedia}, 24:\penalty0 2648--2661, 2021.

\bibitem[Xie et~al.(2024)Xie, Meng, Zhao, Nguyen, Yang, and Zheng]{xie2024dynamic}
Jianyang Xie, Yanda Meng, Yitian Zhao, Anh Nguyen, Xiaoyun Yang, and Yalin Zheng.
\newblock Dynamic semantic-based spatial graph convolution network for skeleton-based human action recognition.
\newblock In \emph{Proceedings of the AAAI Conference on Artificial Intelligence}, pages 6225--6233, 2024.

\bibitem[Xu and Takano(2021)]{Xu2021}
Tianhan Xu and Wataru Takano.
\newblock Graph stacked hourglass networks for 3d human pose estimation.
\newblock In \emph{2021 IEEE/CVF Conference on Computer Vision and Pattern Recognition (CVPR)}, pages 16100--16109, 2021.

\bibitem[Yan et~al.(2018{\natexlab{a}})Yan, Xiong, and Lin]{SpatialYan2018}
Sijie Yan, Yuanjun Xiong, and Dahua Lin.
\newblock Spatial temporal graph convolutional networks for skeleton-based action recognition.
\newblock In \emph{Proceedings of the Thirty-Second AAAI Conference on Artificial Intelligence and Thirtieth Innovative Applications of Artificial Intelligence Conference and Eighth AAAI Symposium on Educational Advances in Artificial Intelligence}. AAAI Press, 2018{\natexlab{a}}.

\bibitem[Yan et~al.(2018{\natexlab{b}})Yan, Xiong, and Lin]{yan2018spatial}
Sijie Yan, Yuanjun Xiong, and Dahua Lin.
\newblock Spatial temporal graph convolutional networks for skeleton-based action recognition.
\newblock In \emph{Proceedings of the AAAI conference on artificial intelligence}, 2018{\natexlab{b}}.

\bibitem[Yang et~al.(2021{\natexlab{a}})Yang, Wang, Dantcheva, Garattoni, Francesca, and Bremond]{yang2021unik}
Di Yang, Yaohui Wang, Antitza Dantcheva, Lorenzo Garattoni, Gianpiero Francesca, and Francois Bremond.
\newblock Unik: A unified framework for real-world skeleton-based action recognition.
\newblock \emph{BMVC}, 2021{\natexlab{a}}.

\bibitem[Yang et~al.(2021{\natexlab{b}})Yang, Liu, Lu, Er, and Kot]{yang2021skeleton}
Siyuan Yang, Jun Liu, Shijian Lu, Meng~Hwa Er, and Alex~C Kot.
\newblock Skeleton cloud colorization for unsupervised 3d action representation learning.
\newblock In \emph{Proceedings of the IEEE/CVF International Conference on Computer Vision}, pages 13423--13433, 2021{\natexlab{b}}.

\bibitem[Yang et~al.(2024)Yang, Liu, Lu, Hwa, and Kot]{yang2024one}
Siyuan Yang, Jun Liu, Shijian Lu, Er~Meng Hwa, and Alex~C Kot.
\newblock One-shot action recognition via multi-scale spatial-temporal skeleton matching.
\newblock \emph{IEEE Transactions on Pattern Analysis and Machine Intelligence}, 2024.

\bibitem[Yu et~al.(2023)Yu, Zhao, Fang, Deng, Su, Wang, Gan, Lu, and Wu]{yu2023regularity}
Shoubin Yu, Zhongyin Zhao, Haoshu Fang, Andong Deng, Haisheng Su, Dongliang Wang, Weihao Gan, Cewu Lu, and Wei Wu.
\newblock Regularity learning via explicit distribution modeling for skeletal video anomaly detection.
\newblock \emph{IEEE Transactions on Circuits and Systems for Video Technology}, 2023.

\bibitem[Zaheer et~al.(2022)Zaheer, Mahmood, Khan, Segu, Yu, and Lee]{zaheer2022generative}
M~Zaigham Zaheer, Arif Mahmood, M~Haris Khan, Mattia Segu, Fisher Yu, and Seung-Ik Lee.
\newblock Generative cooperative learning for unsupervised video anomaly detection.
\newblock In \emph{Proceedings of the IEEE/CVF conference on computer vision and pattern recognition}, pages 14744--14754, 2022.

\bibitem[Zeng et~al.(2021)Zeng, Jiang, Ding, Li, Hao, and Qiu]{zeng2021hierarchical}
Xianlin Zeng, Yalong Jiang, Wenrui Ding, Hongguang Li, Yafeng Hao, and Zifeng Qiu.
\newblock A hierarchical spatio-temporal graph convolutional neural network for anomaly detection in videos.
\newblock \emph{IEEE Transactions on Circuits and Systems for Video Technology}, 33\penalty0 (1):\penalty0 200--212, 2021.

\bibitem[Zhang et~al.(2022)Zhang, Tu, Yang, Chen, and Yuan]{PoseSeq2Seq}
Jinlu Zhang, Zhigang Tu, Jianyu Yang, Yujin Chen, and Junsong Yuan.
\newblock Mixste: Seq2seq mixed spatio-temporal encoder for 3d human pose estimation in video.
\newblock In \emph{2022 IEEE/CVF Conference on Computer Vision and Pattern Recognition (CVPR)}, pages 13222--13232, 2022.

\bibitem[Zhang et~al.(2017)Zhang, Lan, Xing, Zeng, Xue, and Zheng]{zhang2017view}
Pengfei Zhang, Cuiling Lan, Junliang Xing, Wenjun Zeng, Jianru Xue, and Nanning Zheng.
\newblock View adaptive recurrent neural networks for high performance human action recognition from skeleton data.
\newblock In \emph{Proceedings of the IEEE international conference on computer vision}, pages 2117--2126, 2017.

\bibitem[Zheng et~al.(2021)Zheng, Zhu, Mendieta, Yang, Chen, and Ding]{PoseFormer}
Ce Zheng, Sijie Zhu, Matias Mendieta, Taojiannan Yang, Chen Chen, and Zhengming Ding.
\newblock 3d human pose estimation with spatial and temporal transformers.
\newblock \emph{Proceedings of the IEEE International Conference on Computer Vision (ICCV)}, 2021.

\bibitem[Zhou et~al.(2017)Zhou, Huang, Sun, Xue, and Wei]{Zhou2017}
Xingyi Zhou, Qixing Huang, Xiao Sun, Xiangyang Xue, and Yichen Wei.
\newblock Towards 3d human pose estimation in the wild: A weakly-supervised approach.
\newblock In \emph{2017 IEEE International Conference on Computer Vision (ICCV)}, pages 398--407, 2017.

\bibitem[Zhu et~al.(2023)Zhu, Ke, Gong, and Bailey]{zhu2023adaptive}
Anqi Zhu, Qiuhong Ke, Mingming Gong, and James Bailey.
\newblock Adaptive local-component-aware graph convolutional network for one-shot skeleton-based action recognition.
\newblock In \emph{Proceedings of the IEEE/CVF Winter Conference on Applications of Computer Vision}, pages 6038--6047, 2023.

\end{thebibliography}
}

\clearpage
\setcounter{page}{1}
\maketitlesupplementary
\section{Experimental Setup Details}
\subsection{Hyperparameters for DT1: Action Classification}
\label{appx:dt1}
As discussed in Section~\ref{sec:dt1}, we evaluate MoFM on NTU-RGB+D~\cite{ntu60Paper} (60 classes) and NTU-RGB+D-120~\cite{ntu120paper} (120 classes) for action recognition. The detailed hyperparameters for training on both datasets are presented in Table~\ref{tab:training_setup_ntu}.

\begin{table}[h]
    \centering
    \caption{Fine-tuning setup details for action classification on NTU RGB+D \cite{ntu60Paper} and  NTU-RGB+D-120 \cite{ntu120paper} datasets.}
    \begin{tabular}{l c c}
        \toprule
        \toprule
        \textbf{Parameter}            & \textbf{NTU-60} & \textbf{NTU-120}  \\ \hline
        Epoch                         & 16              & 20                 \\ 
        Batch Size                    & \multicolumn{2}{c}{24}              \\ 
        Optimizer                     & \multicolumn{2}{c}{AdamW}           \\ 
        Adam \(\beta_1\), \(\beta_2\) & \multicolumn{2}{c}{0.9, 0.999}      \\ 
        Weight Decay                  & \multicolumn{2}{c}{3e-4}            \\ 
        Base Learning Rate            & \multicolumn{2}{c}{1e-8}            \\ 
        Peak Learning Rate            & \multicolumn{2}{c}{1e-4}            \\ 
        Warmup Period                 & \multicolumn{2}{c}{5 epochs}        \\ 
        Learning Rate Decay           & \multicolumn{2}{c}{Cosine}          \\ 
        Dropout                       & \multicolumn{2}{c}{0.4}             \\ \bottomrule
    \end{tabular}
    \label{tab:training_setup_ntu}
\end{table}

\subsection{Hyperparameters for DT2: One-shot Action Calssification}
\label{appx:dt2}
We evaluate MoFM for one-shot action recognition on NTU-RGB+D-120~\cite{ntu120paper}. The model is trained on the auxiliary set (100 samples per class) using supervised contrastive learning~\cite{khosla2020supervised}, where \(m=2\) samples per class are randomly selected in each batch. Training hyperparameters are summarized in Table~\ref{tab:dt2_training_setup}.
\begin{table}[h]
    \centering
    \caption{Fine-tuning setup details for one-shot action classification on NTU-RGB+D-120 \cite{ntu120paper} dataset.} 
    \begin{tabular}{l c}
        \toprule
        \toprule
        \textbf{Parameter}            & \textbf{Value}                \\ \hline
        Epoch                         & 5                            \\ 
        Batch Size                    & 16                             \\ 
        Optimizer                     & AdamW                         \\ 
        Adam \(\beta_1\), \(\beta_2\) & 0.9, 0.999                    \\ 
        Weight Decay                  & 3e-4                          \\ 
        Base Learning Rate             & 1.e-8                        \\ 
        Peak Learning Rate             & 1.e-5                        \\ 
        Warmup Period (Epochs)        & 0.3                       \\ 
        Learning Rate Decay           & Cosine                        \\ 
        Dropout                       & 0.4                           \\ 
        Contrastive Temp              & 0.1                            \\ \bottomrule
    \end{tabular}
    \label{tab:dt2_training_setup}
\end{table}

\subsection{Hyperparameters for DT3: Self-supervised Human Anomaly Detection}
\label{appx:dt3}
As discussed in Section~\ref{sec:sht}, we evaluate MoFM on the SHT~\cite{liu2018future} dataset and its human-centric subset HR-SHT~\cite{morais2019learning} for anomaly detection. The training hyperparameters for both datasets are presented in Table~\ref{tab:fine_tuning_setup_dt3}. Please note that the hyperparameters remain the same for HR-SHT, as it is a subset of the SHT dataset. HR-SHT \cite{morais2019learning} uses the same training set but features a different test set, which focuses exclusively on human-related anomalies.
\begin{table}[h]
    \centering
    \caption{Fine-tuning setup details for self-supervised anomaly detection on SHT \cite{liu2018future} dataset.}
    \begin{tabular}{l c}
        \toprule
        \toprule
        Epoch                         & 1                            \\ 
        Batch Size                    & 16                             \\ 
        Optimizer                     & Adam                        \\ 
        Adam \(\beta_1\), \(\beta_2\) & 0.9, 0.999                    \\ 
        Weight Decay                  & 6e-4                          \\ 
        Base Learning Rate             & 1e-8                        \\ 
        Peak Learning Rate             & 6e-4                        \\ 
        Learning Rate Decay           & OneCycleLR                        \\ 
        Dropout                       & 0.4                           \\ \bottomrule
    \end{tabular}
    \label{tab:fine_tuning_setup_dt3}
\end{table}

\subsection{Hyperparameters for DT4: Supervised Human Anomaly Detection}
\label{appx:dt4}
As discussed in Section~\ref{sec:sup_anomaly}, for supervised anomaly detection we utilized the UBnormal dataset \cite{Acsintoae_CVPR_2022}. Detailed hyperparameters for the fine-tuning process are provided in Table \ref{tab:extended_fine_tuning_setup_dt4} to ensure reproducibility.
\begin{table}[h]
    \centering
    \caption{Fine-tuning setup details for supervised anomaly detection on UBnormal \cite{Acsintoae_CVPR_2022} dataset.}
    \begin{tabular}{l c}
        \toprule
        \toprule
        \textbf{Parameter}            & \textbf{Value}                \\ \hline
        Epoch                         & 8                            \\ 
        Batch Size                    & 16                            \\ 
        Optimizer                     & Adam                         \\ 
        Adam \(\beta_1\), \(\beta_2\) & 0.9, 0.999                    \\ 
        Weight Decay                  & 3e-4                          \\ 
        Base Learning Rate             & 1e-8                        \\ 
        Peak Learning Rate             & 1e-4                        \\ 
        Learning Rate Decay           & OneCycleLR \\ 
        Dropout                       & 0.4                           \\ \bottomrule
    \end{tabular}
    \label{tab:extended_fine_tuning_setup_dt4}
\end{table}

\end{document}